\pdfoutput=1
\documentclass[10pt,logo,copyright]{nvidiatechreport}
\usepackage[authoryear,sort&compress,square]{natbib}
\usepackage{hyperref}
\usepackage{url}
\usepackage{graphicx}
\usepackage{booktabs} 
\usepackage{multirow}
\usepackage{multicol}
\usepackage{xspace}
\usepackage{subcaption}
\usepackage{dsfont}  % indicator function
\usepackage{minted}
\usepackage{pythonhighlight}
\usepackage{slashbox}  % diagonal line in table  % please import slashbox.sty as well.
\usepackage{amsmath,amsfonts,bm}

\def\Tableref#1{Table~\ref{#1}}

\def\Figref#1{Figure~\ref{#1}}

\def\secref#1{section~\ref{#1}}
\def\Secref#1{Section~\ref{#1}}
\def\eqref#1{eq.~\ref{#1}}
\def\Eqref#1{Eq.~\ref{#1}}
\def\1{\bm{1}}

\DeclareMathAlphabet{\mathsfit}{\encodingdefault}{\sfdefault}{m}{sl}
\SetMathAlphabet{\mathsfit}{bold}{\encodingdefault}{\sfdefault}{bx}{n}

\newcommand{\nickname}{MV-SAM\xspace}

\newcommand{\revision}[1]{#1}

\newcommand{\noindentbold}[1]{\noindent\textbf{#1.}}

\newcommand{\picube}{$\pi^{3}$\xspace}
\newcommand{\SAMVideoName}{SAM2-Video\xspace}

\newcommand{\set}[1]{\{#1\}}
\newcommand{\Real}{\mathbb{R}}
\newcommand{\bmu}{\mbox{\boldmath $\mu$}} % checked for ICLR
\newcommand{\bsigma}{\mbox{\boldmath $\sigma$}} % checked for ICLR

\makeatletter
\DeclareRobustCommand\onedot{\futurelet\@let@token\@onedot}
\def\@onedot{\ifx\@let@token.\else.\null\fi\xspace}

\usepackage{wrapfig}

\usepackage[nameinlink]{cleveref}
\crefname{equation}{Eq.}{Eqs.}
\crefname{figure}{Fig.}{Figs.}
\crefname{section}{Sec.}{Sec.}
\crefname{appendix}{App.}{App.}
\crefname{table}{Tab.}{Tabs.}
\crefname{algorithm}{Alg.}{Alg.}
\crefname{thm}{Thm}{Thm}
\Crefname{thm}{Thm}{Thm}
\crefname{prop}{Prop}{Prop}
\crefname{line}{Line}{Lines}
\usepackage{longtable}

\let\cite\citep  % nvidia format recommends to use \citep{...} instead of \cite{...}. So, I override the command here not to change the original text.

\newcommand{\crefnames}[3]{%
  \@for\next:=#1\do{%
    \expandafter\crefname\expandafter{\next}{#2}{#3}%
  }%
}

\title{MV-SAM: Multi-view Promptable Segmentation using Pointmap Guidance}

\author{
Yoonwoo Jeong$^{1,2,\dagger}$, Cheng Sun$^1$, Yu-Chiang Frank Wang$^1$, Minsu Cho$^2$, Jaesung Choe$^1$\\
$^1$NVIDIA, $^2$POSTECH\\
}

\begin{abstract}

Promptable segmentation has emerged as a powerful paradigm in computer vision, enabling users to guide models in parsing complex scenes with prompts such as clicks, boxes, or textual cues. 
Recent advances, exemplified by the Segment Anything Model (SAM), have extended this paradigm to videos and multi-view images. 
However, the lack of 3D awareness often leads to inconsistent results, necessitating costly per-scene optimization to enforce 3D consistency. 
In this work, we introduce MV-SAM, a framework for multi-view segmentation that achieves 3D consistency using pointmaps--3D points reconstructed from unposed images by recent visual geometry models. 
Leveraging the pixel–point one-to-one correspondence of pointmaps, MV-SAM lifts images and prompts into 3D space, eliminating the need for explicit 3D networks or annotated 3D data. 
Specifically, MV-SAM extends SAM by lifting image embeddings from its pretrained encoder into 3D point embeddings, which are decoded by a transformer using cross-attention with 3D prompt embeddings. 
This design aligns 2D interactions with 3D geometry, enabling the model to implicitly learn consistent masks across views through 3D positional embeddings. 
Trained on the SA-1B dataset, our method generalizes well across domains, outperforming SAM2-Video and achieving comparable performance with per-scene optimization baselines on NVOS, SPIn-NeRF, ScanNet++, uCo3D, and DL3DV benchmarks. 
Code will be released.
\end{abstract}

\begin{document}
\maketitle
\abscontent

\section{Introduction}
\label{sec:introduction}

Promptable segmentation has emerged as a cornerstone of computer vision, enabling humans to efficiently guide models in parsing complex scenes through simple prompts such as clicks, boxes, or textual cues. 
The success of the Segment Anything Model (SAM)~\citep{SAM} has demonstrated the potential of this paradigm across a wide range of 2D applications. 
Building on SAM, follow-up works~\citep{cheng2022xmem, bekuzarov2023xmempp, cheng2023putting} extend promptable segmentation to videos and multi-view images to achieve consistent scene-level segmentation. 
Most notably, SAM2~\citep{SAM2} introduces a mask propagation mechanism based on memory-attention layers, achieving strong cross-domain generalization.
Despite these advances, methods relying on temporal continuity inherently lack 3D awareness, making them prone to occlusion, object reappearance, or repetitive visual patterns. 

A key challenge in incorporating 3D awareness into segmentation models lies in designing a 3D representation that can naturally operate with 2D user prompts.
In detail, conventional 3D representations--such as point clouds, meshes, voxel grids, or 3D Gaussians~\citep{3dgs}--decouple 3D geometry from 2D imagery.
Consequently, mapping 2D user prompts (e.g., a mouse click on an image) into the corresponding 3D position typically requires rendering or projection steps, which introduce unnecessary complexity and remain susceptible to occlusion.

Recent advances in visual geometry models~\citep{wang2025vggt,wang2025pi3} provide a new opportunity. 
Visual Geometry Grounded Transformer (VGGT)~\citep{wang2025vggt} and its follow-up work \picube~\citep{wang2025pi3} have demonstrated that dense 3D reconstructions can be obtained directly from unposed images in a single feed-forward manner while bypassing conventional structure-from-motion pipelines~\citep{colmap1,colmap2}. 
Resulting reconstructions, termed \textit{pointmap}, are geometrically faithful and preserve a strict one-to-one correspondence between pixels and points.
% Each 3D point corresponds to exactly one pixel, while occlusion handling is implicitly resolved. 
This property forms a natural bridge between 2D prompts and 3D geometry, eliminating the need for rendering or projection.

\begin{figure}[!t]
    \centering
    \includegraphics[width=\linewidth,]{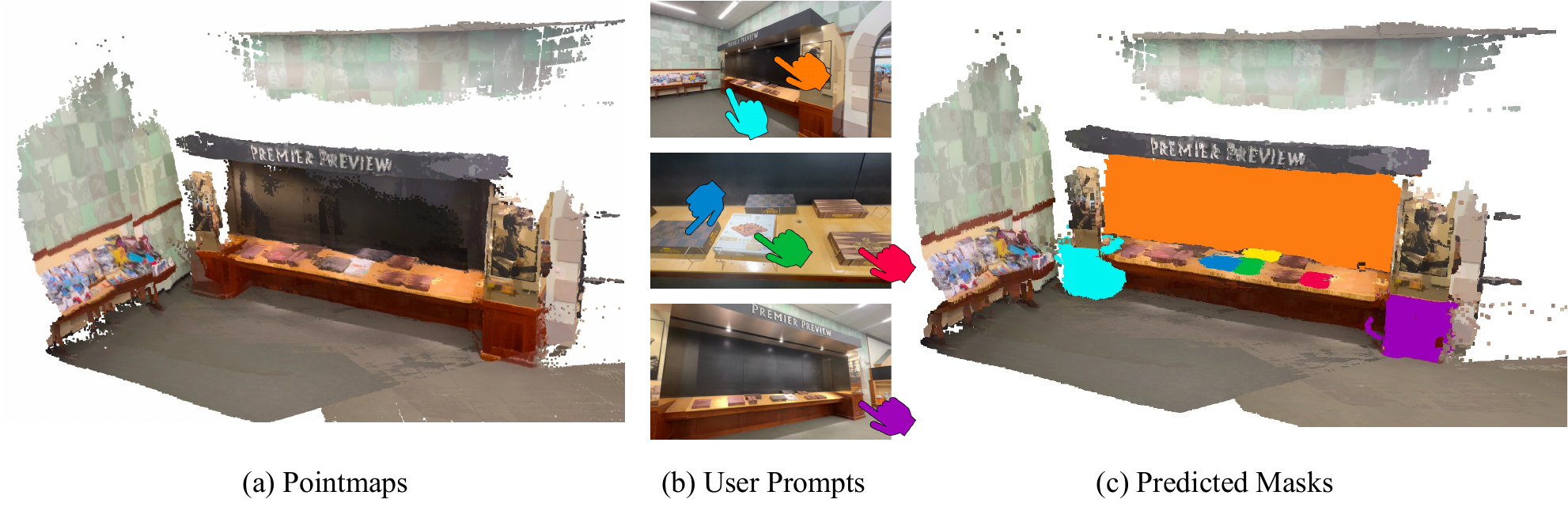}
    \includegraphics[width=\linewidth,]{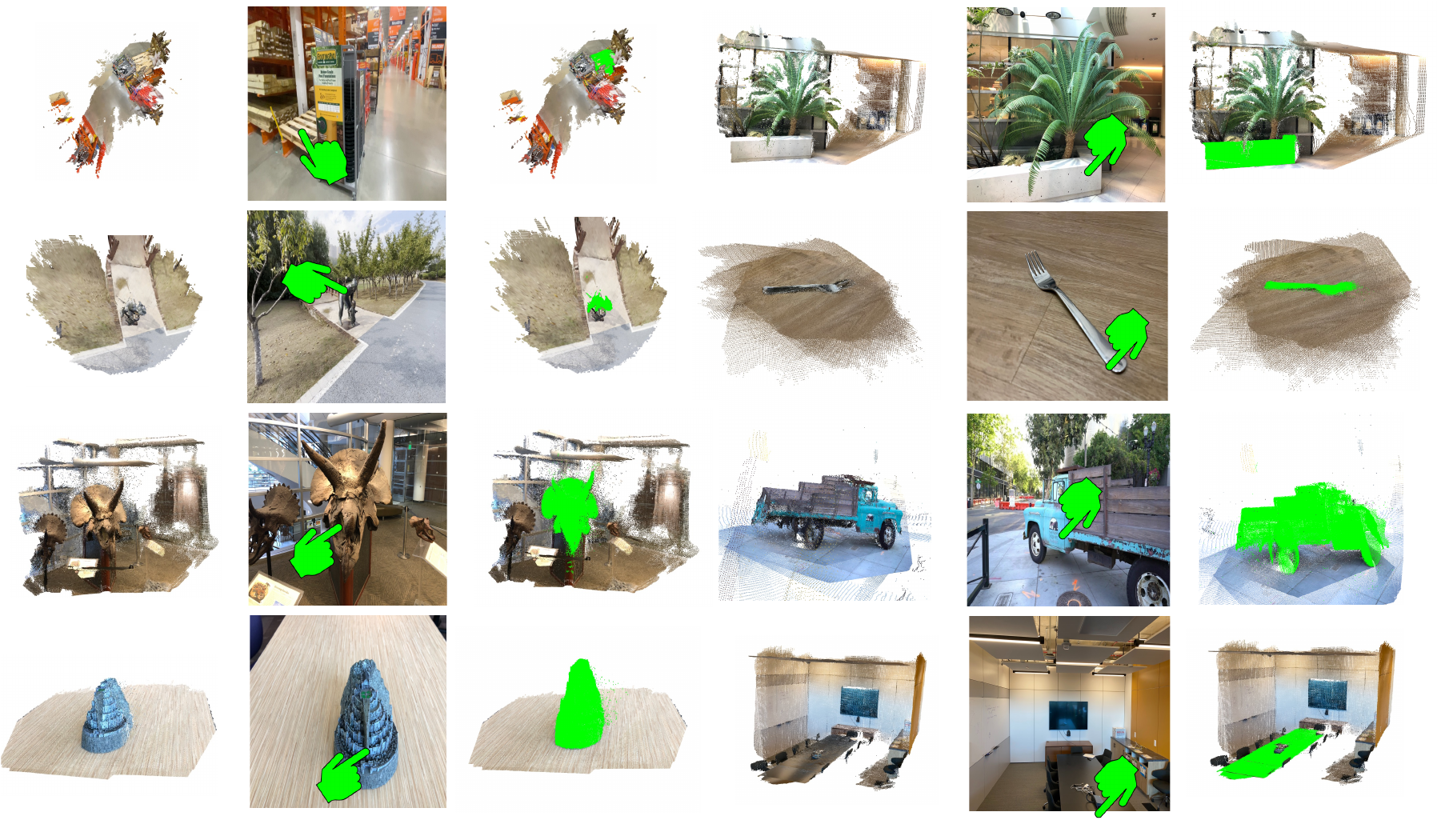}
    % Try not to change the size of this figure since I fit this figure and introduction section within page 2.
    \vspace{-6mm}
    \caption{\nickname. Our method enables view-consistent promptable segmentation, where user prompts (e.g., clicks) guide the extraction of target masks across multi-view images. The examples above illustrate visualizations of the lifted predicted masks via pointmaps, where finger points (illustrating user prompts) share the same color as the corresponding predicted masks. }
    \vspace{-6mm}
    \label{fig:teaser}
\end{figure}

Building on this insight, we propose \textbf{\nickname}, a multi-view promptable segmentation framework that operates on pointmaps reconstructed from unposed images by the visual geometry model. By leveraging the strict one-to-one pixel-point correspondence, we seamlessly transfer pretrained 2D segmentation knowledge (e.g., from SAM) into 3D space without relying on specialized 3D architectures or large-scale annotated 3D datasets. Our framework is threefold:
(1) reusing the \SAMVideoName's image encoder to extract rich image embeddings,
(2) embedding 3D positions on both image embeddings and prompts,
and (3) decoding masks using a lightweight transformer network.
 % that perform cross-attention between point embeddings and 3D prompt embeddings.

% Our framework consists of two components: (1) reusing \SAMVideoName's image encoder to extract rich image embeddings and lifting them into 3D point embeddings, and (2) applying 3D positional embeddings into a simple transformer-based mask decoder that computes cross-view and cross-pixel attention to integrate geometry with user interactions.

We evaluate \nickname on various benchmarks: NVOS~\citep{NVOS}, SpIn-NeRF~\citep{spinnerf}, ScanNet++~\citep{yeshwanth2023scannetpp}, uCo3D~\citep{liu2025uco3d}, and DL3DV~\citep{ling2024dl3dv10k} as visualized in~\Figref{fig:teaser}. 
Our \nickname consistently outperforms the image/video-based foundation model \SAMVideoName~\citep{SAM2}, and achieves competitive performance against optimization-based baselines~\citep{SA3D,SA3D-GS,SAGA,OmniSeg3D}.
In summary, our contributions are threefold:
\begin{itemize}
    % \item We propose a 2D-3D promptable segmentation method based on pointmaps--3D points reconstructed from unposed images, avoiding rendering and projection.
    \item We address multi-view promptable segmentation using pointmaps--3D points reconstructed from unposed images--which eliminate the need for rendering or projection to align 2D interactions with 3D geometry.
    \item We leverage the one-to-one pixel–point correspondence of pointmaps to lift both user prompts and image embeddings from the pretrained \SAMVideoName encoder into 3D space, thereby enabling our model to directly transfer rich 2D knowledge into 3D understanding.
    \item We demonstrate that a lightweight transformer network with 3D positional embeddings is sufficient for robust, view-consistent, and 3D-aware promptable segmentation, eliminating the need for specialized 3D networks or mask-annotated 3D data.
\end{itemize}

\section{Related Work}

\noindentbold{Promptable segmentation}  
Promptable segmentation aims to predict binary masks conditioned on both input modality and user inputs such as clicks, boxes, or masks. 
The Segment Anything Model (SAM)~\citep{SAM} popularized this task by introducing a large-scale, prompt-driven image segmenter capable of generalizing across a wide range of domains and capture conditions.
Its extensions incorporate temporal cues to handle videos, either by combining SAM with trackers~\citep{cheng2023tracking, yang2023track, cheng2023segment, rajic2023segment} or by introducing memory mechanisms as in SAM2~\citep{SAM2}. 
While highly effective in 2D, these models lack explicit 3D understanding, which often leads to predict inconsistent masks across views. %  or under occlusions. 
Our work addresses this limitation by directly linking 2D prompts to 3D geometry through pointmaps.
% , addressing the aforementioned issues.
% The Segment Anything Model (SAM)~\citep{SAM} popularized this task by introducing a large-scale, prompt-driven image segmenter that generalizes across diverse objects and capture conditions. 

\noindentbold{Multi-view segmentation}
Multi-view segmentation aims to achieve consistent segmentation across multiple viewpoints.
Prior work often enforces this consistency through explicit 3D data representations or specialized 3D neural networks.
Some approaches optimize volumetric representations with geometric constraints~\citep{NVOS}, while others leverage neural fields~\citep{nerf, 3dgs} to produce consistent multi-view masks~\citep{spinnerf}.
Another line of work lifts 2D segmentations into 3D, either by fusing SAM predictions into pointclouds~\citep{SA3D} or by learning contrastive 3D feature spaces~\citep{OmniSeg3D}.
More recently, 3D Gaussian Splatting method~\citep{3dgs} have been explored for interactive and promptable segmentation, where Gaussians are augmented with features~\citep{Click-Gaussian}, affinities~\citep{SAGA}, or propagation strategies~\citep{SAGonline}.
While effective, these methods rely on heavy per-scene optimization or computationally demanding pipelines.
In parallel, SAMPro3D~\citep{xu2025sampro3d} introduces 3D promptable segmentation by leveraging 2D–3D correspondences, but its reliance on pre-existing 3D data limits applicability to diverse domains.
In contrast, our framework achieves strong generalization without explicit 3D supervision or scene-specific optimization.

\noindentbold{3D reconstruction from unposed images}  
Reconstructing 3D structures from images has long been a central challenge in computer vision. 
Traditional Structure-from-Motion~\citep{colmap1,colmap2,pan2024glomap} relies on bundle adjustment and iterative optimization, thereby requiring extensive optimization time. 
To improve efficiency, recent studies have proposed feed-forward predictors of pointmaps~\citep{wang2024dust3r, leroy2024mast3r}, which were subsequently extended with global alignment strategies~\citep{wang2024spann3r, wang2025cut3r}.
Recent work such as FASt3R~\citep{Yang_2025_Fast3R} and FLARE~\citep{Zhang_2025_FLARE} introduced feed-forward architectures for jointly predicting camera poses and 3D points. 
VGGT~\citep{wang2025vggt} advanced this line by fully predicting dense pointmaps, camera poses, depth, and point-tracking features, demonstrating strong cross-domain generalization. 
Building on this, \picube~\citep{wang2025pi3} employs permutation-equivariant transformers to address order sensitivity, enabling robust reconstructions in both static and dynamic scenes. 
In this work, we leverage the pointmap predictor \picube to estimate 3D positions of image pixels and user prompts, enabling 2D prompts to be seamlessly transferred into 3D space.

\noindentbold{3D network architecture}
\revision{
To process sparse 3D data, 3D vision researchers have explored network architectures capable of operating on unordered point sets. 
As pioneers in this direction, PointNet~\citep{pointnet} and PointNet++~\citep{qi2017pointnet++} introduce MLP-based architectures that directly consume a fixed number of points. 
However, applying pure MLPs to scene-level point clouds leads to prohibitive memory consumption. 
Consequently, follow-up work such as MinkowskiNet~\citep{choy2019minkowski} proposes memory-efficient sparse convolutions on voxelized representations. 
More recently, inspired by the success of transformers in various vision domains, PTv3~\citep{wu2024ptv3} introduces a transformer-based architecture that achieves state-of-the-art performance across multiple benchmarks.
}

\revision{
Despite these advances, contemporary 3D architectures still rely heavily on metric-depth aligned point clouds and often struggle to generalize to unseen domains. 
In this work, we demonstrate that removing explicit 3D operations from MV-SAM leads to significantly stronger generalization performance.
}

\section{Methodology}
\label{sec:Methodology}

% 3D promptable segmentation task aims to predict 3D masks from pointclouds. While pointclouds can be measured in various ways, but in this paper, we focus on pointmaps--3D points reconstructed from recent visual geometry models. 
% \yw{ promptable segmentation task aims to predict 3D masks from point clouds. While the 3D points can be obtained through various ways, in this work we focus on pointmaps--3D points reconstructed by recent visual geometry models.}

Multi-view promptable segmentation aims to localize all corresponding regions from unposed images of a scene using seed user prompts on the images.  
Let us denote the collection of $N$ unposed images by $\mathcal{I} = \{I_i \}_{i=1}^{N}$ and the set of seed prompts by $\mathcal{S} = \{S_i \}_{i=1}^{N_\text{S}}$ given to $N_\text{S}$ images among them ($N_\text{S} < N$) such that the input can be divided into a prompted set $\mathcal{I_\text{S}} = \{ (I_i, S_i) \}_{i=1}^{N_\text{S}}$ and a unprompted set  $\mathcal{I_\text{U}} = \{ I_i \}_{i=N_\text{S}+1}^N$.
Then, the task is defined to predict the set of object/part masks $\hat{\mathcal{M}} = \set{\hat{M}_i}_{i=1}^{N}$ for all the images, corresponding to the given prompts $\mathcal{S}$. 
% Note that here we only consider positive point prompts (as in SAM2~\citep{SAM2} ), which point to the target regions.

% \yw{
% Multi-view promptable segmentation aims to produce view-consistent masks from unposed images.
% Our framework is defined on triplets $\{(\mathcal{I}, \mathcal{S}, \mathcal{M})\}$, 
% each element consisting of an unposed image, $\mathcal{I}$, with associated user prompts, $\mathcal{S}$, where $\mathcal{S} \in \bigl\{\, \emptyset \,\bigr\} 
% \;\cup\; \bigl\{\, S \mid S \subseteq [0,H) \times [0,W) \,\bigr\}$ with $H$ and $W$ denoting the height and width of $\mathcal{I}$, and the corresponding ground-truth mask $\mathcal{M}$. 
% The problem is formulated as follows: given a set of unposed images with associated user prompts
% $\{(\mathcal{I}, \mathcal{S})\},$ the task is to predict the corresponding set of masks $\{\mathcal{M}\}$ across the viewpoints.
% Note that we only assume positive point prompts from SAM2, which are intended to select the target region.
% }

% These 2D masks are interchangeable to 3D masks~$\set{\mathcal{M}^{\text{PE}}}$ thanks to the property of pointmaps, one-to-one mapping between pixels and points. 
% In~\Figref{fig:overview}, we show predicted masks across various domains using LiftSAM and their corresponding 3D lifts via \picube.
% By doing so, our method can predict both 2D and 3D masks simultaneously as visualized in~\Figref{fig:overview}.

\begin{figure}[!t]
    \centering
    \vspace{-8mm}
    \includegraphics[width=\linewidth]{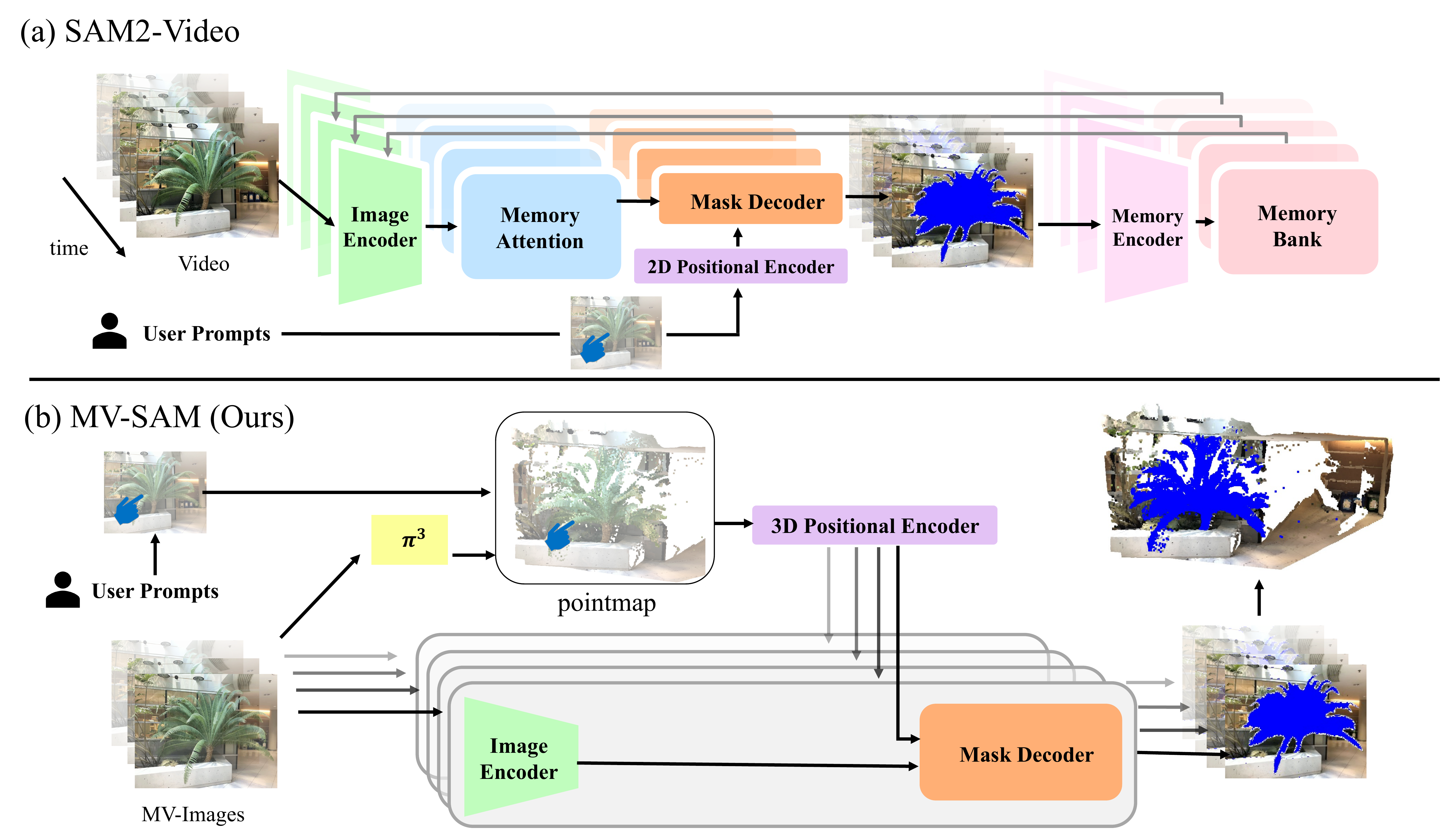}
    \vspace{-6mm}
    \caption{
        Overview. (a) \SAMVideoName tracks masks iterative through memory modules where the visual cues are the key for tracking. In contrast, (b) our \nickname leverage pointmap as a unified world coordinate and embed 3D positional information in both image embeddings and user prompts to predict view-consistent masks without 3D-specific networks or large-scale 3D annotated datasets. 
        % \jchoe{Draw 3D points.} \jchoe{don't forget Prof. Cho's comments.}
        % Overview. (a) \SAMVideoName tracks masks iteratively through memor0y modules, which struggles with visually repetitive objects and often fails to ensure multi-view consistency. In contrast, (b) our \nickname leverages 3D positional embeddings in a unified world coordinate system using pointmaps from \picube~\citep{wang2025pi3}. This design induces view-consistent predictions without 3D-specific neural networks or large-scale 3D annotated datasets. 
        %% Jimmy
        % \SAMVideoName iteratively tracks masks through memory modules, which is subject to error accumulation and fails to guarantee multi-view consistency. In contrast, \nickname performs joint mask prediction across viewpoints by leveraging 3D prompts and pointmaps derived from \picube~\citep{wang2025pi3}. By consistently defining prompts across views, \nickname attains view-consistent predictions without relying on specialized 3D network architectures, while inherently enforcing 3D consistency through the use of 3D positions and prompts.
    }
    % \vspace{-6mm}
\label{fig:overview}
\end{figure}

In~\Figref{fig:overview}, we provide an overview of our MV-SAM. Our framework consists of three stages:
(1) a pre-processing stage that reconstructs pointmaps from unposed images using \picube~\citep{wang2025pi3} and extracts image embeddings using the pretrained image encoder of \SAMVideoName~\citep{SAM2} (\Secref{subsec:Pre-processing stage}),
(2) a positional embedding stage that locates image embeddings and user seed prompts on 3D and embed 3D positional information(\Secref{subsec:3DPE for Prompts and VP}),
and (3) a mask decoding stage that predicts view-consistent masks (\Secref{subsec:Mask decoder}).

\subsection{Pre-processing stage}
\label{subsec:Pre-processing stage}

In this stage, we process a set of unposed images~$\mathcal{I} = \set{I_i}_{i=1}^N$ to obtain pointmaps~$\mathcal{P} = \set{P_i}_{i=1}^N$ and corresponding image embeddings~$\mathcal{F} = \set{F_i}_{i=1}^N$.
We begin by applying an off-the-shelf visual geometry model, \picube~\citep{wang2025pi3}, to reconstruct a pointmap $P_i = [ \mathbf{p}_{ip} ]_{p=1}^{N_\text{P}}$, where $N_\text{P}$ denotes the number of points in the pointmap and each 3D point is represented as $\mathbf{p}_{ip}
\in \Real^3$.
\revision{For detailed preliminaries about \picube, please refer to Appendix A.2.}
The point~$\mathbf{p}$ has strict one-to-one correspondence with an image pixel~$\mathbf{r}_{ip}
\in \Real^{2}$, eliminating the need for rendering or projection to bridge 2D and 3D data. In addition to 3D coordinates, the \picube predicts a set of confidence maps $\mathcal{C}=\set{C_i}_{i=1}^N$, where each confidence map~$C_i$ consists of the value~$c_{ip} \in \Real$ that indicates the reconstruction reliability of corresponding $\mathbf{p}_{ip}$. 
During this stage, we also obtain 3D prompts, $\{S_i^{\text{3D}}\}_{i=1}^{N_s}$, by mapping $\mathcal{S}$ into their corresponding 3D points using $\mathcal{P}$.

\subsection{3D Positional Embedding for User Prompts and Image embeddings}
\label{subsec:3DPE for Prompts and VP}

Previous promptable segmentation models interpret user prompts within images or points to predict the indicated regions. 
Among them, \SAMVideoName (\Figref{fig:overview}a) relies on 2D positional embeddings, assigned independently to each frame, together with memory modules that propagate masks and prompts across frames. 
For multi-view promptable segmentation (\Figref{fig:overview}b), we instead embed both user prompts and image features with 3D positional encodings, defined in a common world coordinate system derived from pointmaps~$\mathcal{P}$, to enable consistent mask prediction across views.

\noindentbold{3D positional embeddings from pointmaps}
We observe that the extracted pointmaps exhibit varying standard deviations depending on the number of frames, which brings various scene scales.
To address this, we empirically found that applying standardization across all points yields consistent predictions that remain robust under changes in the number of frames. 
For empirical details, please refer to  Appendix~\ref{subsec:standard}.
Specifically, we compute the mean and standard deviation over all points~$\mathcal{P}$ and apply z-score normalization to enforce a unit standard deviation and zero mean.
The standardized coordinates~$\Tilde{\mathbf{p}}_{ip} \in \Tilde{P}_i$ are then passed through the sinusoidal positional embedding computed as:

% \begin{equation}\label{eq:3dpe}
%     \mathcal{F}^{\text{PE}} = [\sin(2\pi\mathbf{b}^\top \Tilde{\mathcal{P}}),~ \cos(2\pi\mathbf{b}^\top \Tilde{\mathcal{P}})]^\top~~s.t.~~\Tilde{\mathcal{P}} = \frac{\mathcal{P} - \mu}{\sigma},
% \end{equation}
\begin{equation}\label{eq:3dpe}
    \mathbf{f}^{\text{PE}}_{ip} = [\sin(2\pi\mathbf{b}^\top \Tilde{\mathbf{p}}_{ip}),~ \cos(2\pi\mathbf{b}^\top \Tilde{\mathbf{p}}_{ip})]^\top~~~\text{where}~~~\Tilde{\mathbf{p}}_{ip} = \frac{\mathbf{p}_{ip} - \bmu}{\bsigma},
\end{equation}
where $\bmu \in \Real^3$ is the mean, 
$\bsigma \in \Real^3$ is the standard deviation, 
% $\Tilde{{P}}$ is the standardized points,
$\mathbf{f}^{\text{PE}}_{ip} \in F_i^{\text{PE}}$ is a positional embedding vector, and $\mathbf{b} \in \Real^{3\times 64}$ is the Fourier basis frequency~\citep{sam_pe,SAM}.
% We proceed with this process across the entire points such that we obtain a set of 3D positional embeddings~${F}^{\text{PE}} = \set{F^{\text{PE}}_i}_{i=1}^N$ where each pointmap~$P$ has corresponding 3D positional embedding~$F^{\text{PE}} = \set{\mathbf{f}^{\text{PE}}_j}_{j=1}^{N_\mathcal{P}}$ where $N_\mathcal{P}$ is the number of all points.

\noindentbold{Prompt embeddings}
We mostly follow the prompt encoder design introduced by SAM~\citep{SAM} except for the positional embeddings. We use 3D positional embeddings instead of 2D. 
% Given the seed prompts~$\mathcal{S}$, we index out the corresponding 3D positional embeddings from $\mathcal{F}^{\text{PE}}$.
Then the formulation to compute prompt embedding from a 3D prompt $\mathbf{s}_{ip}^{\text{3D}} \in S_i^{\text{3D}}$ is:
% \begin{equation}
%     \mathcal{S}^{\text{PE}} =
%     \begin{cases}
%     \mathcal{F}^{\text{PE}}[\mathcal{S}], & \mathcal{S} \neq \emptyset, \\
%     \emptyset, & \mathcal{S} = \emptyset,
%     \end{cases}
% \end{equation}
\begin{equation}
    \mathbf{s}^{\text{PE}}_{ip} =
    \begin{cases}
    \mathbf{f}_{ip}^{\text{PE}} + \mathbf{f}^{\text{pos}}, & i \in [1, N_\text{S}]~~\text{and}~~\mathbf{s}^{\text{3D}}_{ip}~~\text{is a positive prompt},  \\
    \mathbf{f}_{ip}^{\text{PE}} + \mathbf{f}^{\text{neg}}, & i \in [1, N_\text{S}]~~\text{and}~~\mathbf{s}^{\text{3D}}_{ip}~~\text{is a negative prompt}, 
    % \
    % \mathbf{0}, & \text{else},
    \end{cases}
\end{equation}
where $\mathbf{f}^{\text{pos}}\in\Real^{128}$ and $\mathbf{f}^{\text{neg}}\in\Real^{128}$ are learnable embeddings corresponding to positive and negative prompt, respectively. 
% where $[\cdot]$ denotes the index operator.
% that extracts the corresponding~$\mathcal{F}^{\text{PE}}$.
% By doing so, we obtain a set of point embeddings~$\mathcal{F}^{\mathcal{P}} = \set{F^{\mathcal{P}}_i}_{i=1}^N$.

% Given pointmaps, we apply standardization on 3D locations and use the sinusoid positional embedding introduced by NeRF~\citep{nerf}. We empirically found that there is notable performance improvement with this normalization as stated in~\Tableref{table:ablation_mask_decoder}. 
% \chengs{(Not clear how the standardization actually work. Min-max scale to fit into [0, 1] for nerf pe?)} \jchoe{I realized that standardization and normalization are distinct terms, as noted on Wikipedia. I recall that Jimmy applied standardization--that is, subtracting the mean and dividing by the standard deviation. As Cheng suggested, it would be helpful to include some additional explanation here.}

\noindentbold{Confidence embeddings}
While \picube generally provides accurate reconstructions, points with low confidence can still be inaccurately localized, and incorporating such points as prompts adversely affects segmentation performance.
To mitigate this issue, we introduce two learnable embeddings: one for high-confidence points and another for low-confidence points.
These embeddings are added to the positional embeddings of both user prompts and pointmaps, allowing the model to modulate its attention according to the confidence score maps~$\mathcal{C}$.
We define low-confidence points as the bottom \revision{15\%} of all points ranked by confidence, denoted by $c^{\text{th}}$. For each pixel coordinate $\mathbf{r}$ and each seed prompt $\mathbf{s}$, we then define their positional embeddings as follows::
\begin{equation}\label{eq:confidence_embed}
    \hat{\mathbf{f}}_{ip}^{\text{PE}} = 
    \begin{cases}
        \mathbf{f}_{ip}^{\text{PE}} + \mathbf{f}^{\text{hc}},~~c_{ip} > c^{\text{th}}, \\
        \mathbf{f}_{ip}^{\text{PE}} + \mathbf{f}^{\text{lc}},~~\text{else},
    \end{cases}~~~~
    \hat{\mathbf{s}}_{ip}^{\text{PE}} = 
    \begin{cases}
        \mathbf{s}_{ip}^{\text{PE}} + \mathbf{f}^{\text{hc}},~~c_{ip} > c^{\text{th}}, \\
        \mathbf{s}_{ip}^{\text{PE}} + \mathbf{f}^{\text{lc}},~~\text{else},
    \end{cases}
\end{equation}
where $\mathbf{f}^{\text{hc}}\in\Real^{128}$ and $\mathbf{f}^{\text{lc}}\in\Real^{128}$ are learnable embeddings corresponding to high-confidence and low-confidence score, respectively. 
$c^{\text{th}} \in \Real$ is a threshold for the confidence. 
In our experiments, we dynamically set the threshold $c^{\text{th}}$ by selecting the lowest \revision{15}\% of confidence scores across all views.

% $\mathbf{r}\in\Real^2$ are image pixels and $\mathbf{s}\in\Real^2$ are each pixel location of the seed prompt.

\noindentbold{Point embeddings}
Given an image embedding vector~$\mathbf{f}_{ip} \in F_i$ and a 3D positional embedding vector~$\hat{\mathbf{f}}_{ip}^{\text{PE}} \in \hat{F}_i^{\text{PE}}$, we calculate a point embedding~$\hat{\mathbf{f}}_{ip}^{\text{PE}} \in \hat{F}_{i}^{\text{PE}}$~~by~~$\hat{\mathbf{f}}_{ip}^{\text{PE}} = \mathbf{f}_{ip} + \hat{\mathbf{f}}_{ip}^{\text{PE}}$.

% In the meantime, \picube generally provides accurate reconstructions, points with low confidence can still be inaccurately localized. Incorporating such points as prompts adversely affect segmentation performance.
% To address this, we introduce two learnable embeddings: one for high-confidence points and the other low-confidence points. These embeddings are added to the positional embeddings of both user prompts and pointmap points, enabling the model to adapt its attention according to confidence score maps~$\mathcal{C}$. 
% For threshold, we use the low 20\% confidence from all points.

% Following the decoder architecture of SAM2, we add the processed 3D positional embeddings to the image embeddings before each attention layer in the mask decoder. Note that we keep the same prompt encoder as in \SAMVideoName, but replace the 2D positional embeddings with 3D positional embeddings.

\subsection{Mask decoder}
\label{subsec:Mask decoder}

Given point embeddings $\hat{\mathcal{F}}^{\mathcal{P}}$ and prompt embeddings~$\hat{\mathcal{S}}^{\text{PE}}$, our mask decoder is trained to predict masks~$\hat{\mathcal{M}} = \set{\hat{M}_i}_{i=1}^N$. 
\revision{
We adopt the two-way transformer design of \SAMVideoName, which restricts the attention scope to individual viewpoints. 
Specifically, \SAMVideoName applies independent 2D positional embeddings to each frame and uses memory attention to implicitly track masks from previous frames. 
In contrast, our mask decoder utilizes 3D positional embeddings to place all image pixels within a unified world coordinate system defined by pointmaps. 
The primary difference between the two methods is that \SAMVideoName propagates predicted masks to subsequent frames via a memory attention layer, whereas MV-SAM inherently propagates prompts across all frames using 3D positional embeddings.
}
Accordingly, our mask decoder is able to locate all the possible prompts~$\hat{\mathcal{S}}^{\text{PE}}$ for every~$\hat{F}^{\mathcal{P}}_i$ through transformer layers to predict masks~$\hat{M}_i$ (referred to as `single-view' in~\Tableref{table:ablation_mask_decoder}). 
Specifically, our decoder takes $\hat{F}^{\mathcal{P}}_i$ as queries and $\hat{\mathcal{S}}^{\text{PE}}$ as keys and values, and the output mask $\hat{M}_i$ is formulated as:
% Consequently, our transformer layers process each viewpoint independently while sharing the 3D prompt embeddings across all views. For each viewpoint, our decoder takes $\hat{F}^{\mathcal{P}}_i$ as queries and $\hat{\mathcal{S}}^{\text{PE}}$ as keys and values, and the output mask $\hat{M}_i$ is formulated as:
\begin{equation}\label{eq:decoder}
\hat{M}_i = \text{Decoder}(\hat{F}^{\mathcal{P}}_i,~\hat{\mathcal{S}}^{\text{PE}}).
\end{equation}
% where the mask decoder is to locate all the possible prompts~$\hat{\mathcal{S}}^{\text{PE}}$ for every~$\hat{F}^{\mathcal{P}}_i$ through transformer layers to predict masks~$\hat{M}_i$ (referred to as `single-view' in~\Tableref{table:ablation_mask_decoder}). 
% that computes masks by computing view-wise attention as~$\hat{\mathcal{M}} = \text{Decoder}(\hat{\mathcal{F}}^{\mathcal{P}},~\hat{\mathcal{S}}^{\text{PE}})$ (referred to as `full-view' in~\Tableref{table:ablation_mask_decoder}). 

Our architecture is fully equivariant to frame order, since \picube is designed to satisfy permutation equivariance. This property ensures that the model yields consistent performance even when the input frames are randomly permuted, as shown in~\Tableref{tab:sam2-ours-prompts}.

\subsection{Training loss}
\label{subsec:Training loss}
We train our model using a standard binary segmentation loss that supervises the predicted masks against ground truth object masks. For scalability, we train our model on SA-1B~\citep{SAM}, a large-scale dataset with mask annotations of objects. 
Importantly, our model is trained solely on single-view object–image pairs, without requiring any multi-view datasets to achieve multi-view consistency.
Specifically, we employ a combination of focal loss~\citep{lin2017focal} and dice loss~\citep{milletari2016vnet} to handle class imbalance and ensure sharp mask boundaries.

Given ground-truth 2D masks~$\mathcal{M}$, we first randomly sample sparse or dense prompts from~$\mathcal{M}$. Then, our method infers predicted masks~$\hat{\mathcal{M}}$ from unposed images~$\mathcal{I}$ and the prompts~$\mathcal{S}$ through the frozen image encoder~$\theta_{\mathrm{imgenc}}^*$, trainable parameters~$\theta$ such as the mask decoder~$\theta_{\mathrm{dec}}$, the prompt encoder~$\theta_{\mathrm{penc}}$, and the confidence embeddings~$\theta_{\mathrm{conf}}$, which is optimized by minimizing the losses:
\begin{equation}
    \min_{\theta} \mathcal{L} = \min_{\theta} \big( \lambda_{\mathrm{focal}}\mathcal{L}_{\mathrm{focal}} + \lambda_{\mathrm{dice}}\mathcal{L}_{\mathrm{dice}} \big),
\end{equation}
Detailed hyperparameters are described in Section~\ref{subsec:implementation_details} of the Appendix.

% which is formulated as:
% \begin{equation}
%     \hat{\mathcal{M}} = \text{MV-SAM}\big(\mathcal{I}, \mathcal{S} \mid \theta_{\mathrm{imgenc}}^*, \theta_{\mathrm{dec}}, \theta_{\mathrm{promptenc}}, \theta_{\mathrm{conf}}
%     \big).
% \end{equation}

% Then, our network is trained to minimize the segmentation loss~$\mathcal{L}$ following objectives:
% \begin{equation}
%     \mathcal{L} = \lambda_{\mathrm{focal}}\mathcal{L}_{\mathrm{focal}} + \lambda_{\mathrm{dice}}\mathcal{L}_{\mathrm{dice}},
% \end{equation}
% where detailed hyperparameters are described in Section~\ref{subsec:implementation_details} of the supplementary material.

\section{Experiments}
\label{sec:experiments}

\begin{table}[!t]
\centering
\caption{
    Comparison of \SAMVideoName and our \nickname under different prompt settings using three benchmarks: ScanNet++, uCo3D, and DL3DV. \revision{Note that we have slightly improved the performance of MV-SAM by training for more epochs without modifying any hyperparameters. In addition, we refined the manually annotated masks in DL3DV, which resulted in updated performance.
    }
}
\vspace{-2mm}
\resizebox{0.95\linewidth}{!}{
\begin{tabular}{ccccc}
\toprule

\multirow{3}{*}{Dataset} 
& \multicolumn{2}{c}{Video} 
& \multicolumn{2}{c}{Multi-view Images} \\

\cmidrule(lr){2-3} \cmidrule(lr){4-5}
& \SAMVideoName & \nickname & \SAMVideoName & \nickname \\

& mIoU ($\uparrow$) / mAcc ($\uparrow$) & mIoU ($\uparrow$) / mAcc ($\uparrow$) & mIoU ($\uparrow$) / mAcc ($\uparrow$) & mIoU ($\uparrow$) / mAcc ($\uparrow$) \\
\midrule

ScanNet++ & 46.1 / 61.4 & \textbf{\revision{48.9}} / \textbf{\revision{63.5}} & 47.5 / 62.8 & \textbf{\revision{49.1}} / \textbf{\revision{62.9}} \\
uCo3D & 81.9 / 91.3 & \textbf{\revision{87.7}} / \textbf{\revision{95.0}} & 83.2 / 91.9 & \textbf{\revision{87.4}} / \textbf{\revision{95.1}} \\
DL3DV & \revision{67.3} / \revision{82.9} & \revision{\textbf{75.1}} / \revision{\textbf{91.8}} & \revision{64.2 / 78.6} & \revision{\textbf{75.0} / \textbf{92.0}} \\
\midrule
Average & \revision{65.1 / 78.5} & \revision{\textbf{70.6 / 83.4}} & \revision{65.0 / 77.8} & \revision{\textbf{70.5 / 83.3}} \\
\bottomrule

\end{tabular}
}
\label{tab:sam2-ours-prompts}
\end{table}
\begin{figure}[!t]
    \centering
    \includegraphics[height=0.5\linewidth]{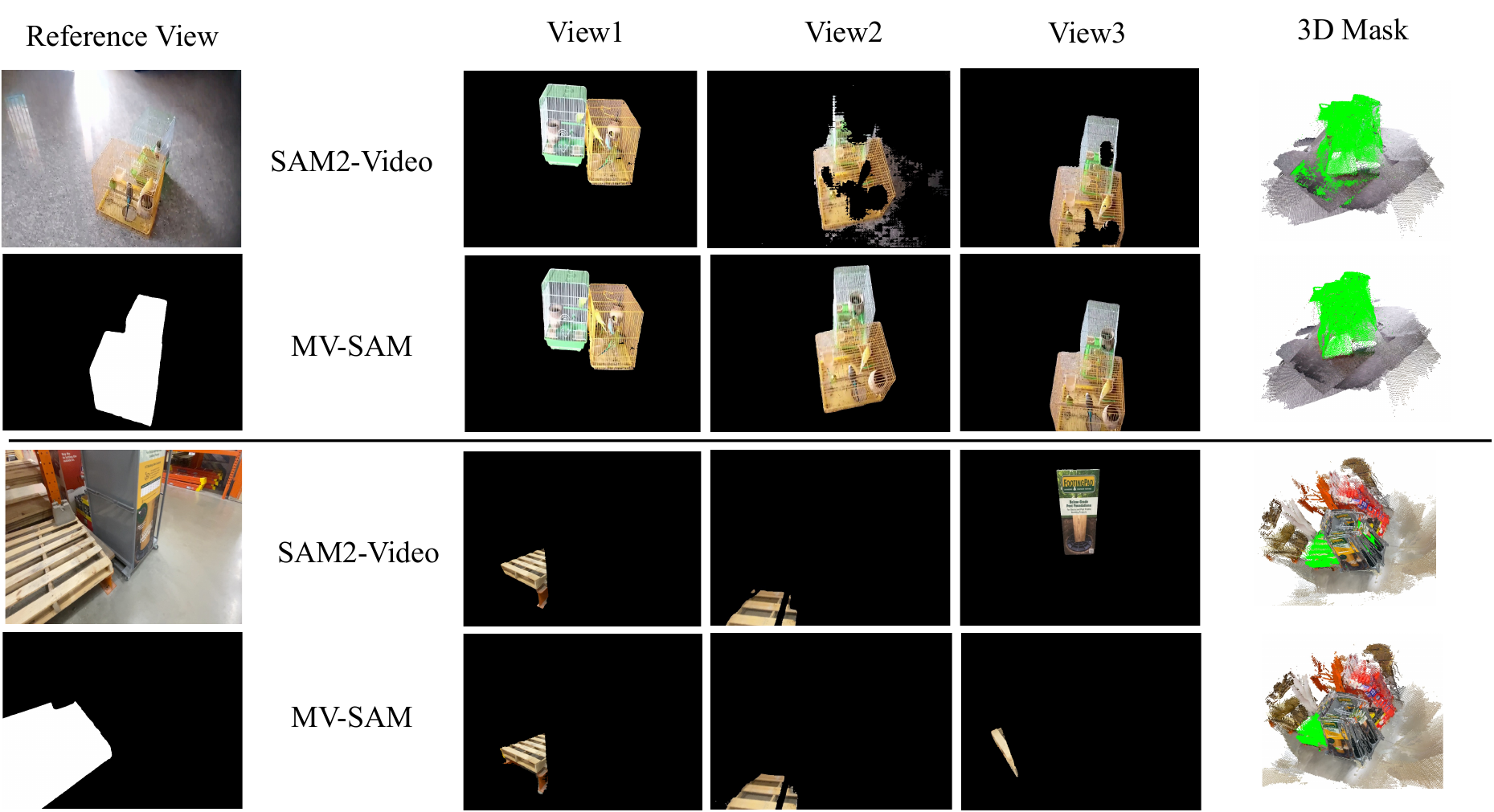}  % change width once Jimmy put the resized figures
    \vspace{-2.5mm}
    \caption{Comparison of \nickname with \SAMVideoName. }
    \label{fig:main_sam_comparison}
    \vspace{-3.5mm}
\end{figure}

%%%%%%%%%%%%%%%%%%%%%%%%%%%%%%%%%%%%%%%%%%%%%%%%%%%%%%
% I relocate this subsection in the appendix due to page limit.
%%%%%%%%%%%%%%%%%%%%%%%%%%%%%%%%%%%%%%%%%%%%%%%%%%%%%%
% \subsection{Implementation Details}
% We train our model on SA-1B~\citep{SAM} dataset that contains 1B images with multiple object masks per image.
% For each image, we sample 10 prompts, allowing up to 10\% to be negative.
% To supervise dense prompts, we randomly drop 80\% of the ground-truth mask and perturb 20\% of the initial mask to simulate errors.
% For focal loss, we set $\alpha=0.9$ and $\gamma=1.5$, with loss weights $\lambda_{\text{focal}}=1.0$, $\lambda_{\text{DICE}}=0.05$, and $\lambda_{\text{IoU}}=0.3$.
% As the image encoder, we adopt the publicly available SAM2.1 large checkpoint, while for the mask decoder, we use the \SAMVideoName's image decoder but train it from scratch.
% Since our framework does not employ an object score predictor, we remove components related to stability and object scores.
% To suppress sprinkles, we adopt the sprinkle-removal strategy from \SAMVideoName, discarding regions whose area is smaller than 0.1\% of the total pixels.
% For evaluation, we select between 10 and 100 frames depending on dataset characteristics. 
% % Additional details are provided in the Appendix.

\subsection{Promptable Segmentation on Videos or Multi-view Images}
We compare \nickname with \SAMVideoName across diverse domains to assess the generalizable capability of our model. The evaluation covers multiple benchmarks: real-world object-centric scenes~\citep{liu2025uco3d}, outdoor scenarios~\citep{ling2024dl3dv10k}, and indoor environments~\citep{yeshwanth2023scannetpp}. 
To reflect practical scenarios, we consider two evaluation settings: (1) Video: temporally coherent video inputs, and 
(2) MV-Images: video frames randomly permuted to simulate multi-view image inputs.
For evaluation, we randomly sample 10 positive prompts and 2 negative prompts from the target masks across different viewpoints.
Positive prompts are sampled from within the target region, whereas negative prompts are sampled from non-target regions.
For fairness, we use the same set of prompts when comparing the models.

\Tableref{tab:sam2-ours-prompts} showcases the superiority of \nickname over \SAMVideoName across different domains. 
Our \nickname consistently outperforms SAM2-Video across diverse datasets in both video and multi-view image settings. 
As illustrated in the first example of Figure~\ref{fig:main_sam_comparison}, SAM2-Video often fails to reliably track objects, frequently introducing holes or missing object parts during mask propagation. 
% In the second example of Figure~\ref{fig:main_sam_comparison}, SAM2-Video is shown to track incorrect objects due to its lack of 3D awareness, whereas our \nickname achieves consistent object tracking. 
% Furthermore, since both \picube and our model architecture ensure equivariance under frame permutations, \nickname maintains consistent performance even when the input frame order is randomized. Most importantly, unlike \SAMVideoName, which requires densely annotated videos and thus requires costly manual labeling, \nickname can be trained directly on readily available 2D segmentation datasets, making it significantly more practical.

\subsection{Multi-view Segmentation: NVOS and SpIn-NeRF}
We further evaluate \nickname, \SAMVideoName and prior per-scene optimization methods (NeRF-based approaches~\citep{spinnerf, OmniSeg3D}, Gaussian-based approaches~\citep{SAGA, SA3D-GS}, and depth-based methods~\citep{SA3D}) on the NVOS~\citep{NVOS} and SpIn-NeRF~\citep{spinnerf} benchmarks, which are widely used for evaluating multi-view promptable segmentation performance\footnote{We exclude the `orchid' scene in NVOS and the `pinecone' scene in SpIn-NeRF; see the Appendix~\Secref{subsec:implementation_details} for reasoning and details.}. 

The datasets consist of multi-view images without guaranteed temporal consistency across frames. Following the evaluation protocol of SAGA~\citep{SAGA}, we randomly sample 8 positive prompts and 2 negative prompts. For NVOS, the prompts are derived from scribbles, while for SPIn-NeRF they are sampled from reference-view masks.

\begin{table}[!t]
\centering
\caption{Comparison on NVOS and SPIn-NeRF benchmarks. \revision{Note that we have slightly improved the performance of MV-SAM by training for more epochs without modifying any hyperparameters.}}
\vspace{-2mm}
\resizebox{1.0\linewidth}{!}{
\begin{tabular}{cccccc}
\toprule
\multirow{2}{*}{Category} & \multirow{2}{*}{Method} & \multicolumn{2}{c}{NVOS} & \multicolumn{2}{c}{SPIn-NeRF} \\
\cmidrule(lr){3-4} \cmidrule(lr){5-6}
 & & mIoU ($\uparrow$) & mAcc ($\uparrow$) & mIoU ($\uparrow$) & mAcc ($\uparrow$) \\
\midrule
\multirow{5}{*}{per-scene opt.}
 & SPIn-NeRF~\citep{spinnerf}          & -    & -    & 90.7 & 98.8 \\
 & SA3D~\citep{SA3D}            & 91.1 & 98.4 & 92.4 & 98.8 \\
 & SAGA~\citep{SAGA}            & 92.6 & 98.6 & 93.7 & 99.2 \\
 & SA3D-GS~\citep{SA3D-GS}      & 92.7 & 98.5 & 93.4 & 99.1 \\
 & OmniSeg3D~\citep{OmniSeg3D}  & 92.8 & 98.6 & 94.5 & 99.3 \\
\midrule
\multirow{2}{*}{generalization}
 & \SAMVideoName~\citep{SAM2} & 88.7  & 94.6   & 86.6 & 93.6 \\
 & \nickname (ours) & \revision{\textbf{92.1}}  & \revision{\textbf{97.5}}   & \revision{\textbf{92.9}}    & \revision{\textbf{97.1}} \\
\bottomrule
\end{tabular}
\label{table:nvos_spin}
}
\end{table}

\begin{figure}[!t]
    \centering
    \includegraphics[width=\linewidth]{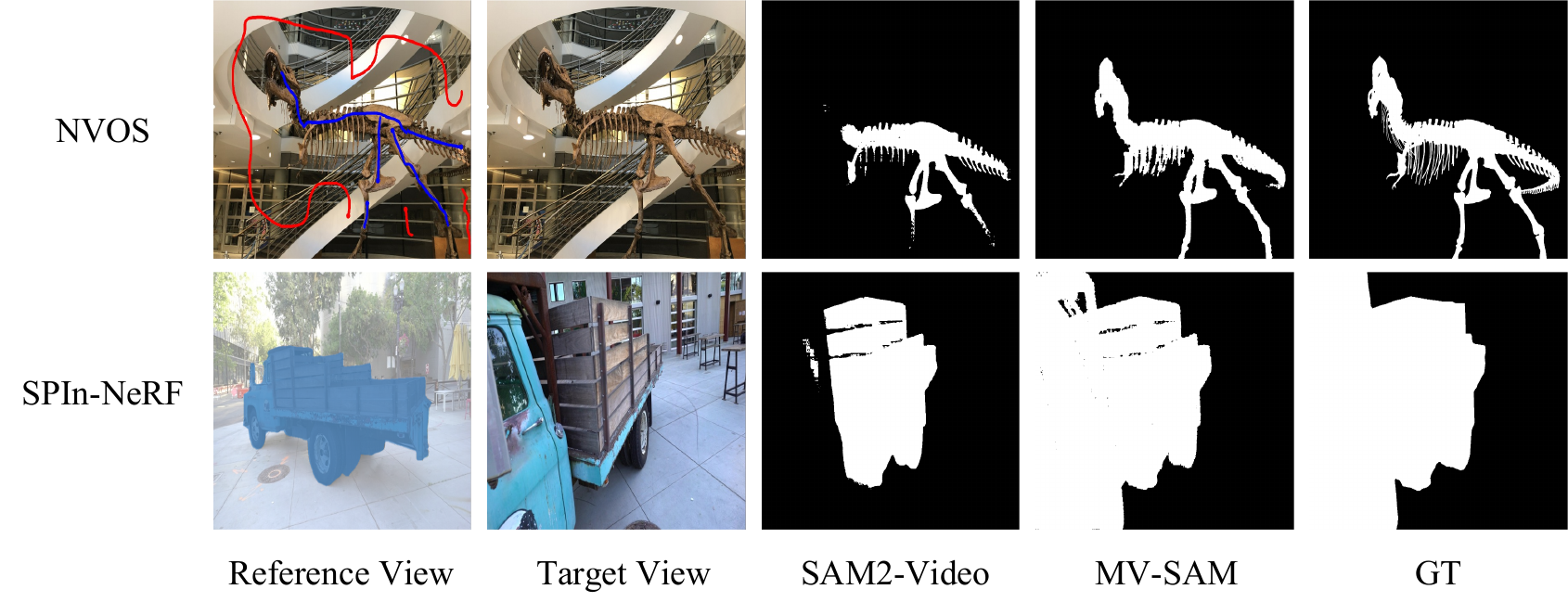}
    \vspace{-6mm}
    \caption{Qualitative results on NVOS and SPIn-NeRF datasets. Compared to SAM2-Video, which often predicts incorrect regions due to the lack of 3D awareness, \nickname achieves consistent mask predictions. In NVOS, user prompts are provided as scribbles, where blue lines visualize positive scribbles and red lines visualize negative scribbles, whereas in SPIn-NeRF, object masks in the reference view serve as user prompts. }
    \label{fig:nvos_spin}
    % \vspace{-4mm}
\end{figure}

As shown in~\Tableref{table:nvos_spin}, our method outperforms \SAMVideoName, while performing competitive with per-scene optimization approaches.
However, the heavy computational cost of per-scene optimization makes those methods impractical for interactive graphics.
It is also worth noting that our model is trained without using any scenes from NVOS or SPIn-NeRF.
In~\Figref{fig:nvos_spin}, SAM2-Video fails to capture the heads of the T-Rex and the trucks, suggesting that the model struggles to segment complete objects when their colors are similar to the background. 
In contrast, our \nickname produces reliable object masks by leveraging the underlying 3D structure, which separates the T-Rex and the trucks from the background in the pointmaps and thereby simplifies the segmentation problem.

\subsection{Control experiments}
\label{subsec:control_experiments}

We conduct control experiments to demonstrate the effectiveness of MV-SAM. Note that our control experiments are performed on the ScanNet++~\citep{yeshwanth2023scannetpp} dataset, which serves as the dataset for both training and evaluation.

\noindentbold{Confidence embeddings}
We assign distinct embeddings to low-confidence points and prompts, enabling the model to recognize which inputs are less reliable as described in~\Eqref{eq:confidence_embed}. 
% Through confidence-guided positional embeddings (CPE), the model explicitly encodes the confidence levels of prompts and pointmap points. 
As shown in~\Tableref{tab:ablation}, incorporating confidence embeddings yields a 7.7$\%p$ improvement, suggesting that, without this guidance, the model becomes susceptible to errors arising from inaccurate geometries of low-confidence points.
Furthermore, we investigate the effect of varying the proportion of low-confidence points considered, with detailed results provided in~\Tableref{tab:CGPE_supp} of the Appendix.

\begin{table}[t]
  \centering
  \caption{
    Ablation study of (a) the mask decoder network and (b) the encoder network. CP denotes the usage of confidence embeddings, PE refers to positional embeddings, Norm indicates normalization of 3D positions as described in~\secref{subsec:Pre-processing stage}, and Attn abbreviates the attention operator. We use MinkUnet42 (37.9M)~\citep{choy2019minkowski} for 3D encoder baseline. 
    `3D rep.’ stands for 3D representation where we introduce residual blocks for 2D-to-3D conversion by adopting sparse voxel representation~\citep{choy2019minkowski} and point representation~\citep{wu2024ptv3}.
  }
  \vspace{-2mm}
  \label{tab:ablation}
  \begin{minipage}{0.424\textwidth}  % 0.413
    \centering
    \subcaption{Mask decoder network.}
    \vspace{-2mm}
    \resizebox{1.00\linewidth}{!}{
    \begin{tabular}{c c c c c}
      \toprule
      CP & PE & Attn & mIoU ($\uparrow$) & mAcc ($\uparrow$) \\
      \midrule
       & 3D PE & single-view & 44.5 & 61.1 \\
      \checkmark & 3D PE & single-view & \textbf{52.2} & \textbf{66.7} \\
      \midrule
      \checkmark & No PE & full-view & 25.6 & 57.8 \\
      \checkmark & No PE & single-view & 10.9 & 52.7 \\
      \checkmark & 2D PE & full-view & 26.6 & 59.5 \\
      \checkmark & 2D PE & single-view & 18.3 & 53.6 \\
      \checkmark & 3D PE & full-view & 45.8 & 62.2 \\
      \checkmark & 3D PE & single-view & \textbf{52.2} & \textbf{66.7} \\
      \bottomrule
    \end{tabular}
    }
    \label{table:ablation_mask_decoder}
  \end{minipage}%
  \hfill
  \begin{minipage}{0.565\textwidth}  % 0.577
    \centering
    \subcaption{Encoder network.}
    \vspace{-2mm}
    \resizebox{1.00\linewidth}{!}{
      \begin{tabular}{c c c c c}
        \toprule
        
        Encoder & 3D rep. & Grid Size & mIoU ($\uparrow$) & mAcc ($\uparrow$) \\
        \midrule

        3D encoder (Mink) & Voxel (Mink) & 0.005 & 37.2 & 64.4 \\
        \midrule
        
        \multirow{3}{*}{Image encoder (SAM2)} & \multirow{3}{*}{Voxel (Mink)} & 0.05 & 40.6 & 63.5 \\
        & & 0.01 & 41.3 & 65.0 \\
        & & 0.005 & 44.3 & 64.5 \\
        \midrule
        
        \multirow{3}{*}{Image encoder (SAM2)} & \multirow{3}{*}{Voxel (PTv3)} & 0.05 & 40.7 & 64.3 \\
        & & 0.01 & 41.0 & 65.2 \\
        & & 0.005 & 42.1 & 66.3 \\
        \midrule
        
        Image encoder (SAM2) & 3DPE (ours) & - &  \textbf{52.2} & \textbf{66.7} \\
        \bottomrule
      \end{tabular}
    }
    \label{table:ablation_encoder}
  \end{minipage}
\end{table}
\begin{figure}[!t]
    \centering    
    \includegraphics[width=1.0\linewidth]{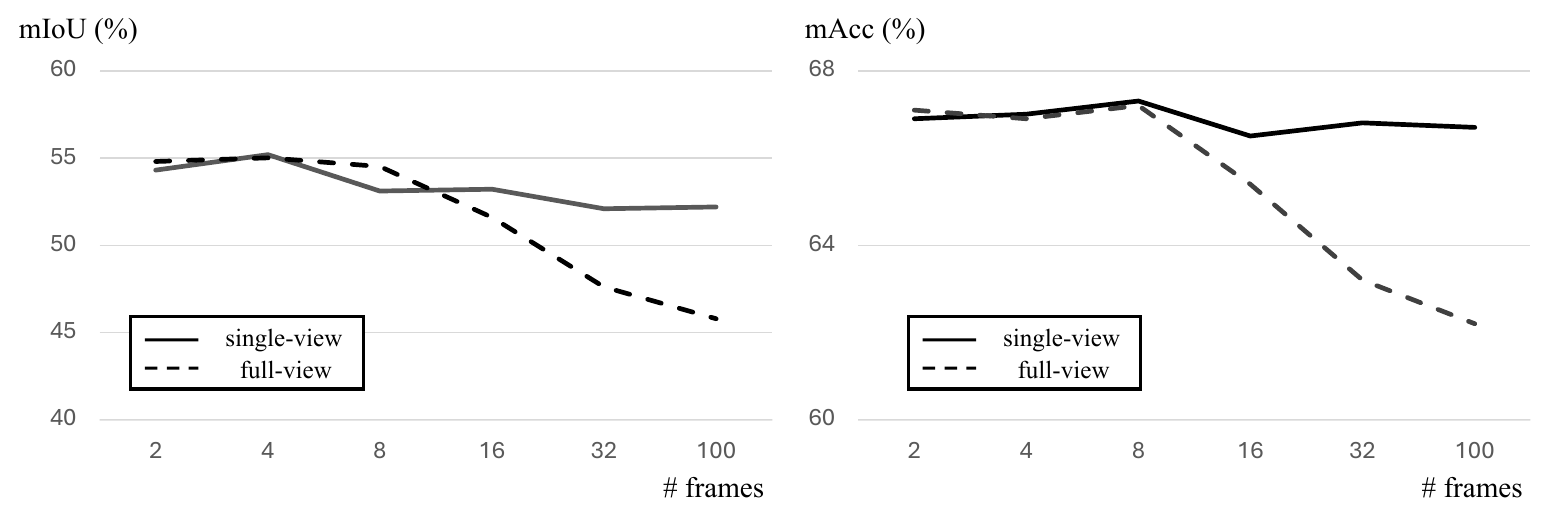}
    \vspace{-9mm}
    \caption{
        Comparison of single-view attention (ours) and full-view attention. While both approaches perform similarly in few-view setups, full-view attention struggles to scale to a larger number of frames, which is common in practice. Detailed results are provided in the Appendix Table~\ref{table:frame_changes}. 
    }
    \label{fig:exp-num-frames}
    % \vspace{-4mm}
\end{figure}

\noindentbold{3D positional embeddings}
In~\Tableref{table:ablation_mask_decoder}, we compare our 3D positional embeddings against the model variants using 2D positional embedding or not using any positional embedding. Without positional embeddings, the model fails to localize prompts and often selects incorrect objects during evaluation. With 2D embeddings, where 3D prompts are projected onto individual image planes, the model struggles to determine which prompts remain valid. In this case, some prompts may correspond to occluded regions or even disappear from the scene, making it difficult for the model to reliably identify the intended target. As a result, the model frequently selects occluding objects instead of the actual object of interest, highlighting the inherent limitations of 2D embeddings in handling occlusion.

\noindentbold{Attention scope}
We compare our single-view attention in the mask decoder with full-view attention that aggregates information across all frames, as shown in~\Tableref{tab:ablation} and~\Figref{fig:exp-num-frames}. 
For a fair comparison, both models are trained with 8 frames per sample. 
While full-view attention achieves performance comparable to single-view attention when the number of frames is limited to 8 or fewer, its performance degrades substantially as the number of frames increases at evaluation. 
This degradation stems from the fact that full-view attention introduces a variable number of tokens depending on the frame count, which may require specialized handling for token length extrapolation~\citep{press2024alibi}, whereas single-view attention maintains a consistent token structure regardless of frame count. 
Therefore, we adopt single-view attention, which avoids the instability caused by token length extrapolation.

\begin{figure}[!t]
    \centering
    \vspace{-4mm}
    \includegraphics[width=1.0\linewidth]{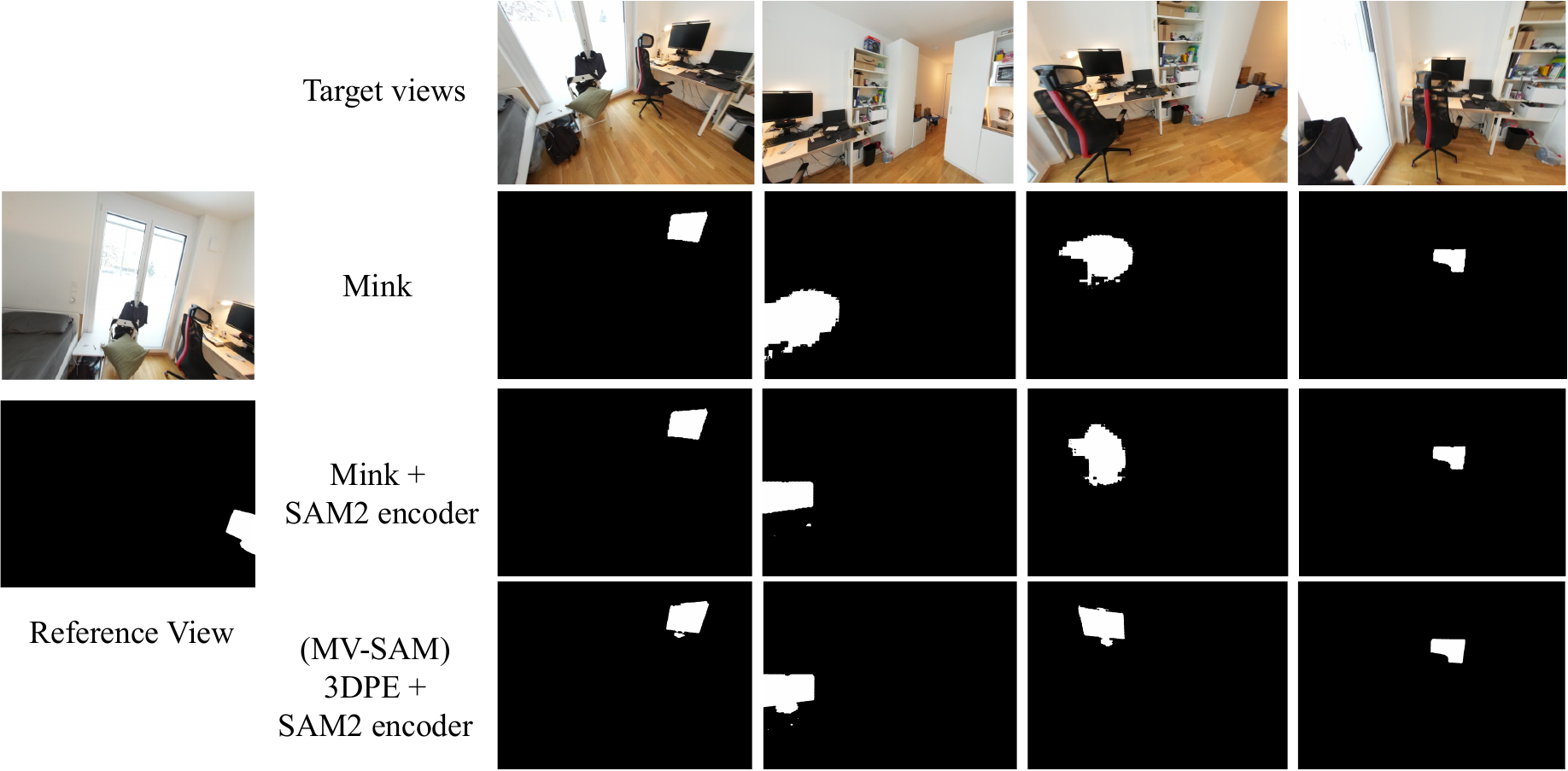}
    \caption{Qualitative results on ablation studies.}
    \label{fig:qual_ablation}
    \vspace{-4mm}
\end{figure}

\noindentbold{Pretrained image encoder from \SAMVideoName}
We also compare our MV-SAM against explicit 3D networks that ensure 3D consistency. 
While keeping the same mask decoder and training losses, we replace our encoder with three variants: 
(1) a pure 3D encoder using MinkUNet, 
(2) a pretrained image encoder with voxel-based residual blocks from MinkUNet~\citep{choy2019minkowski}, 
and (3) from PTv3~\citep{wu2024ptv3}. 
As shown in~\Tableref{table:ablation_encoder}, the pure 3D encoder yields the lowest accuracy, likely because traditional 3D networks assume metric-depth–aligned inputs, whereas pointmaps from visual geometry models often exhibit inconsistent scales. 
Although voxel-based approaches enforce view consistency, their performance is highly sensitive to grid resolution and degrades when pointmap scales vary. 
Moreover, as shown in Figure~\ref{fig:qual_ablation}, the model tends to \revision{produce} blurry rendering due to their restricted resolution by voxel sizes.
In contrast, our model avoids these rigid inductive biases, allowing transformers to implicitly learn 3D consistency--an approach more suitable for pointmaps, which lack metric alignment but preserve rich geometric structure.

\noindentbold{Training on multi-view datasets}
\revision{
A major challenge in training MV-SAM is the scarcity of multi-view mask-annotated datasets.
Although datasets such as ScanNet++~\citep{yeshwanth2023scannetpp} and uCo3D~\citep{liu2025uco3d} are available, they remain limited in scale, restricted in domain, or biased toward single-object scenes. To assess the effect of dataset choice and scale, we compared MV-SAM trained on the small-scale, multi-view datasets (ScanNet++ and uCo3D) against the single-view, large-scale SA-1B dataset as described in Table~\ref{tab:cross_dataset}.
% . The results, summarized in Table \ref{tab:cross_dataset}, clearly illustrate the superior generalization capability provided by large-scale pre-training.
}

\revision{
Models trained on ScanNet++ or uCo3D achieve strong performance when evaluated on their corresponding `in-domain dataset' (e.g., ScanNet++ $\rightarrow$ ScanNet++ achieving 0.510 mIoU). However, their performance drops substantially when tested in a cross-dataset setting, such as the dramatic drop for uCo3D $\rightarrow$ ScanNet++ (0.194 mIoU) and ScanNet++ $\rightarrow$ uCo3D (0.322 mIoU). This confirms that models trained with small-scale, domain-specific multi-view data suffer from poor generalization ability.
}

\revision{
In contrast, the model trained on the single-view, large-scale SA-1B dataset demonstrates consistently strong performance across all domains. When evaluating on ScanNet++, the SA-1B model achieved 0.489 mIoU, nearly matching the in-domain ScanNet++ benchmark and dramatically outperforming the uCo3D cross-domain result. Similarly, SA-1B $\rightarrow$ uCo3D achieves an impressive 0.877 mIoU, almost matching the in-domain uCo3D performance (0.910 mIoU). This evidence highlights the definitive advantage of using a large-scale, diverse dataset like SA-1B as a foundational training resource for MV-SAM, confirming that scale and diversity outweigh multi-view data constraints for achieving robust generalization.
}

% \begin{table}[h]
% \centering
% \resizebox{1.0\linewidth}{!}{
% \begin{tabular}{lcccccc}
% \toprule
% \multirow{2}{*}{\backslashbox{Eval dataset}{Train dataset}} & \multicolumn{2}{c}{SA-1B} & \multicolumn{2}{c}{ScanNet++} & \multicolumn{2}{c}{uCo3D} \\
% \cmidrule(lr){2-3} \cmidrule(lr){4-5} \cmidrule(lr){6-7}
% & mIoU ($\uparrow$) & mAcc ($\uparrow$) & mIoU ($\uparrow$) & mAcc ($\uparrow$) & mIoU ($\uparrow$) & mAcc ($\uparrow$) \\
% \midrule
% ScanNet++ & 0.489 & 0.635 & 0.510 & 0.694 & 0.194 & 0.251 \\
% uCo3D     & 0.877 & 0.950 & 0.322 & 0.517 & 0.910 & 0.965 \\
% \bottomrule
% \end{tabular}
% }
% \caption{Cross-dataset evaluation results.}
% \label{tab:cross_dataset}
% \end{table}

\begin{table}[!t]
\centering
\resizebox{1.0\linewidth}{!}{
\begin{tabular}{lll|cc}
\toprule
Model & Dataset info. & Train dataset $\rightarrow$ Eval dataset & mIoU ($\uparrow$) & mAcc ($\uparrow$) \\
\midrule
\multirow{6}{*}{\nickname (ours)} & In-domain data & ScanNet++ $\rightarrow$ ScanNet++ & 0.510 & 0.694 \\
& Multi-view (small-scale) & uCo3D $\rightarrow$ ScanNet++ & 0.194 & 0.251 \\
& Single-view (large-scale) & SA-1B $\rightarrow$ ScanNet++ & 0.489 & 0.635 \\
\cmidrule(lr){2-5}
& In-domain data & uCo3D $\rightarrow$ uCo3D & 0.910 & 0.965 \\
& Multi-view (small-scale) & ScanNet++ $\rightarrow$ uCo3D & 0.322 & 0.517 \\
& Single-view (large-scale) & SA-1B $\rightarrow$ uCo3D & 0.877 & 0.950 \\
\bottomrule
\end{tabular}
}
\vspace{-2mm}
\caption{Cross-dataset evaluation results.}
\label{tab:cross_dataset}
\vspace{-4mm}
\end{table}
\section{Conclusion}

We presented \textbf{\nickname}, a framework for multi-view promptable segmentation that leverages pointmaps to connect 2D user interactions with multi-view images. By lifting image embeddings from the \SAMVideoName's pretrained image encoder into 3D, our method eliminates the need for explicit 3D networks and achieves view-consistent mask prediction without multi-view supervision or 3D annotated dataset. Extensive experiments across NVOS, SpIn-NeRF, ScanNet++, uCo3D, and DL3DV demonstrate that \nickname consistently outperforms \SAMVideoName and achieves comparable performance with per-scene optimization baselines without any per-scene optimization. 

\revision{
\noindentbold{Limitation} Since our model relies on an off-the-shelf visual geometry model, its performance is inherently tied to the quality of the pointmaps generated by \picube. 
For instance, if \picube produces pointmaps with inaccurate depth alignment or structural noise in highly cluttered indoor scenes, these imperfections can propagate to the downstream segmentation process. 
Furthermore, because our method does not explicitly enforce 3D consistency across views, it may produce unreliable predictions in the presence of outliers, such as misaligned points or artifacts in textureless regions. More description about limitation is included in Appendix H.
}

% While promising, our framework also highlights several directions for future work. First, extending \nickname to dynamic scenes with moving or deformable objects remains an open challenge. Second, applying our approach to domains where explicit 3D geometry is unavailable, such as animations or stylized content, requires further investigation. We believe addressing these limitations will broaden the applicability of promptable 3D segmentation and foster new progress in interactive 3D perception.

\appendix
\section*{Appendix}

\section{Implementation Details}
\label{subsec:implementation_details}
We train our model on SA-1B~\citep{SAM} dataset that contains 1B images with multiple object masks per image.
For each image, we sample 10 prompts, allowing up to 10\% to be negative.
To supervise dense prompts, we randomly drop 80\% of the ground-truth mask and perturb 20\% of the initial mask to simulate errors.
For focal loss, we set $\alpha=0.9$ and $\gamma=1.5$, with loss weights $\lambda_{\text{focal}}=1.0$ and $\lambda_{\text{DICE}}=0.05$.
As the image encoder, we adopt the publicly available SAM2.1 large checkpoint and keeps frozen during training.
\revision{Note that we employ the Hiera-L architecture as the visual backbone, initialized with the pre-trained SAM2 checkpoint (`facebook/sam2.1-hiera-large').}
Since our framework does not employ an object score predictor not mIoU score predictor, we remove components related to stability and object scores.
To suppress sprinkles, we adopt the sprinkle-removal strategy from \SAMVideoName, discarding regions whose area is smaller than 0.1\% of the total pixels.
For evaluation, we select between 10 and 100 frames depending on dataset characteristics. 

\subsection{Dataset}

\begin{figure}[h]
    \centering    
    \includegraphics[width=1.0\linewidth]{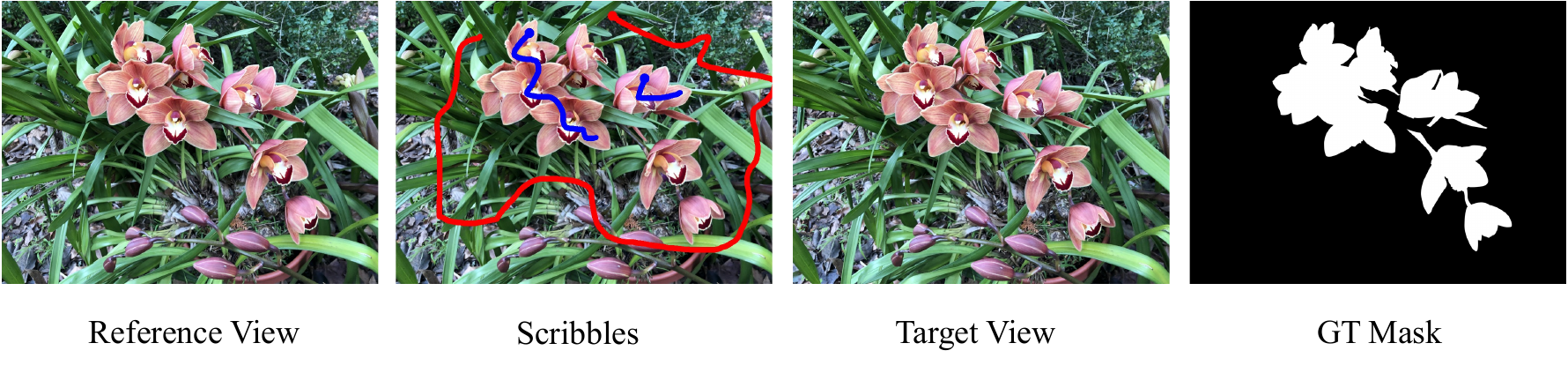}
    \vspace{-6mm}
    \caption{Visualization of the \textit{orchid} scene from the NVOS~\citep{NVOS} dataset.}
    \label{fig:orchids}
\end{figure}

% \paragraph{NVOS.}
% The NVOS
\noindent\textbf{NVOS} dataset~\citep{NVOS} consists of 8 scenes, each with one reference view and one target view. 
We follow the official protocol and use the provided scribble prompts. 
However, the scribble annotation for the \textit{orchid} scene is inaccurate for promptable segmentation. 
As shown in~\Figref{fig:orchids}, the positive prompts cover only 3 out of 7 petals. 
Consequently, SAM2~\citep{SAM2}, SA-3D~\citep{SA3D}, SA3D-GS~\citep{SA3D-GS}, and \nickname yield low average performance, despite correctly segmenting the corresponding petals. 
Therefore, we exclude the \textit{orchid} scene from our evaluation and report the per-scene results of previous baselines and MV-SAM in Section~\ref{subsec:perscene_results}.

% \paragraph{SPIn-NeRF} 
% The SPIn-NeRF
\noindent\textbf{SPIn-NeRF} dataset~\citep{spinnerf} provides multi-view masks for target objects. 
Following the evaluation protocol of SAGA~\citep{SAGA}, we sample all point prompts from the first viewpoint, sorted by file name. 
Since the \textit{pinecone} scene is no longer available from the original data source, it is excluded from our evaluation. 
To facilitate future research, we report the per-scene results, including those of previous baselines, in Section~\ref{subsec:perscene_results}.

% \paragraph{ScanNet++}
% ScanNet++
\noindent\textbf{ScanNet++} dataset~\citep{yeshwanth2023scannetpp} provides 3D instance masks but does not include 2D-level instance annotations. 
The official repository of ScanNet++ offers 2D instance masks rendered via rasterization, which we adopt for evaluating SAM2-Video and \nickname. 
Because ScanNet++ contains objects that occupy only a very small portion of the image, we restrict evaluation to objects covering at least 0.1\% of the total image pixels. 
To ensure coverage of entire scenes, we uniformly sample 100 DSLR frames per scene for evaluation. 
For training and validation, we follow the official ScanNet++ split, and for evaluation, we select 5 objects per validation scene.

% \paragraph{uCo3D}
% uCo3D
\noindent\textbf{uCo3D} dataset~\citep{liu2025uco3d} provides segmentation masks for target objects across large-scale video collections. 
Given the scale of the dataset, we sample 50 representative sequences for our experiments. 
Since uCo3D primarily consists of object-centric videos and does not include wide-baseline scenarios with large viewpoint variations, 
we uniformly sample 50 frames from each sequence to ensure consistent spatial coverage while keeping the evaluation computationally feasible. 
For evaluation, we directly adopt the official masks provided by uCo3D, which allows for a fair comparison with prior work and ensures reproducibility of our results.

% \paragraph{DL3DV}
% DL3DV
\noindent\textbf{DL3DV} dataset~\citep{ling2024dl3dv10k} contains diverse indoor and outdoor scenes, each involving multiple objects. 
Since the dataset does not provide object masks for evaluation, we manually annotated the sequences with the assistance of the SAM2 demo. 
As SAM2 occasionally produced incorrect tracking results, we refined the annotations and re-propagated the masks until reliable masks were obtained. 
Once accurate masks were prepared, we uniformly sampled 100 frames from each sequence, resulting in a total of 5 evaluation samples from DL3DV. 
To ensure transparency and reproducibility, we will release the annotated data together with our code.

\subsection{Model}

\subsubsection{Preliminaries: $\pi^3$}
\revision{
We employ $\pi^3$~\citep{wang2025pi3} as the visual geometry model of MV-SAM. 
Here, we provide a more detailed explanation of $\pi^3$ and its improvements over VGGT~\citep{wang2025vggt}. 
}
\revision{
VGGT is a feed-forward visual geometry transformer that jointly predicts depth, camera poses, and pointmaps from multiple frames. 
A pointmap refers to a dense per-pixel 3D representation in which each image pixel is mapped to a corresponding 3D point in the world coordinate system.
VGGT architecture consists of a DINO-based~\citep{oquab2023dinov2} image tokenization module, a cross-view fusion transformer, and several prediction heads for geometric outputs. 
While effective, VGGT exhibits a critical limitation: its predictions are sensitive to the order of input views. T
his permutation sensitivity arises primarily from sequence-dependent positional encodings and asymmetric cross-view attention, both of which treat the set of input images as an ordered sequence rather than an unordered set. 
As a result, permuting the input frames leads to inconsistent geometric predictions, limiting the robustness of VGGT in the multi-view setup.
}

\revision{
To resolve this limitation, $\pi^3$ introduces a permutation-equivariant reformulation of VGGT’s multi-view interaction mechanism, ensuring that the predicted geometry remains consistent under any permutation of input frames.
Specifically, $\pi^3$ replaces view-index positional encodings with set-based view embeddings that do not encode sequential order, ensuring that each view is treated as an unordered element of a set. 
Additionally, $\pi^3$ adopts a symmetric cross-view attention module in which all views attend to one another with equal structural weight, eliminating the directional bias inherent in VGGT’s attention layers. 
Furthermore, $\pi^3$ incorporates a permutation-invariant geometry aggregation module that fuses multi-view tokens in a manner consistent with set operations. 
These architectural modifications guarantee strict permutation equivariance.
}

\revision{
With these modifications, $\pi^3$ not only resolves the permutation sensitivity observed in VGGT but also substantially improves the quality of geometric predictions. 
The resulting consistency across different view orderings leads to more stable depth and pointmap estimation, ultimately enabling state-of-the-art reconstruction performance on both 3D and 4D benchmarks. 
By adopting $\pi^3$ as its visual geometry model, MV-SAM also inherits permutation equivariance with respect to frame ordering.
}

\subsubsection{Difference between `single-view' and `full-view' attention}
\revision{
In Table~\ref{table:ablation_mask_decoder}, we compare different positional embeddings under various attention strategies. 
Here, we provide additional details on how the single-view and full-view attention mechanisms operate. 
The key distinction lies in the scope of cross-attention between image embeddings and prompt tokens. 
In the single-view setting, cross-attention is computed only between the prompts and the image embeddings of a single reference view. 
In contrast, the full-view setting broadens this interaction by allowing prompts to attend to the image embeddings from all available views. 
For clarity, we additionally provide a PyTorch-like code snippet illustrating both attention modules in Listing~\ref{lst:attention}.
}

% NOTE THAT YOU MUST USE ``2024'' TeX Live version. 
% YOU CAN CHANGE THE SETUP IN THE Settings > Compiler > Tex Live version > 2024
\begin{listing}[t]
\begin{python}
import torch
import torch.nn as nn
from einops import repeat
from jaxtyping import Float

# Transformer network
transformer = nn.Transformer(..., batch_first=True)

# Data
image_embeds: Float[torch.Tensor, "n_views h w d"]
prompt_embeds: Float[torch.Tensor, "n_prompts d"]  # assume click prompt.

# Method 1: single-view attention (i.e., view-wise attention)
query = repeat(prompt_embeds, "n_prompts d -> n_views n_prompts d")
key = image_embeds.reshape(n_views, h * w, d)
out = transformer(query, key)
out = out.reshape(n_views, h, w, d)

# Method 2: full-view attention
query = prompt_embeds.reshape(1, n_prompts, d)
key = image_embeds.reshape(1, n_views * h * w, d)
out = transformer(query, key)
out = out.reshape(n_views, h, w, d)
\end{python}
\vspace{-4mm}
\caption{
Comparison between the single-view and full-view attention modules. The single-view attention has a sequence length of $h \times w$, indicating attention is computed for each individual image. In contrast, the full-view attention uses a sequence length of $\mathrm{n\_views} \times h \times w$, allowing attention to be computed across all different images. Although our mask decoder employs the two-way transformers detailed in SAM2-Video, we have replaced this component with a simple $\operatorname{nn.Transformer}(\cdots)$ in the illustration for ease of comprehension.
% A PyTorch-like code snippet comparing the single-view and full-view attention modules is provided for illustration. In the single-view attention, the sequence length is $h * w$ and it means the attention is computed per individual image. On the other hand, in the full-view attention, the sequence length is $\mathrm{n\_views} * h * w$ which implies the attention is computed across all different images. Note that our mask decoder consists of two-way transformers following SAM2-Video~\cite{SAM2}, however for easy understanding, we simply replace the two-way transformer with simple transformer (nn.Transformer).
% Note that in our actual implementation, we avoid explicit for-loops and instead employ fully batchified operations for efficiency.
}
\label{lst:attention}
\end{listing}

\subsubsection{Reducing Multi-View Prediction to Single-View Training}

\begin{figure}
    \centering
    \includegraphics[width=1.0\linewidth]{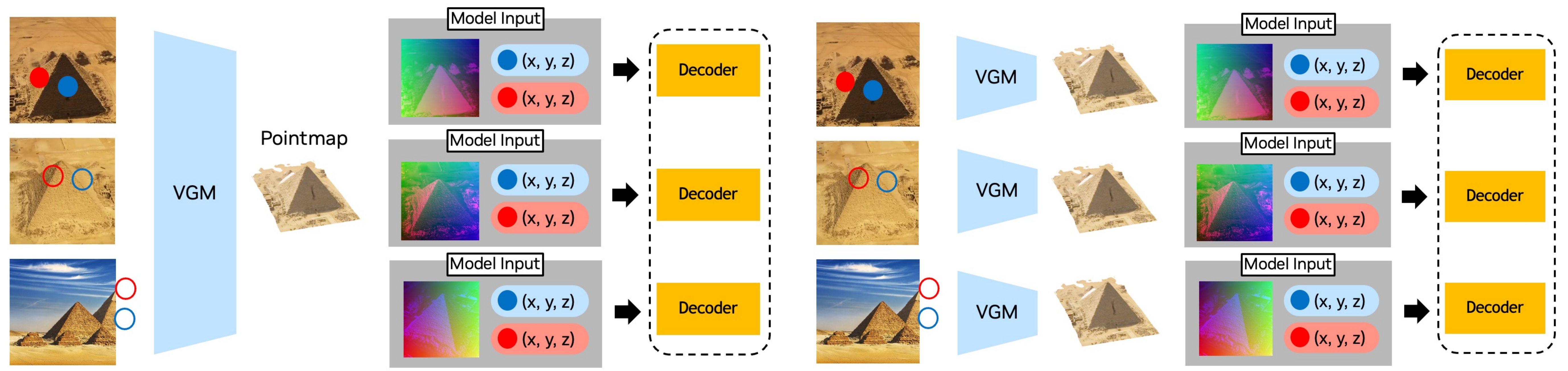}
    \caption{\revision{Illustration of how multi-view prediction can be reduced to a set of independent single-view prediction problems.
    During inference, VGM processes all views jointly (left), whereas during training, each view is handled independently (right).
    Since the VGM produces nearly identical geometric outputs whether frames are processed jointly or individually, the multi-view prediction task can be effectively reformulated as a single-view prediction problem.}}
    \label{fig:multiview}
\end{figure}

\revision{
As illustrated in Figure~\ref{fig:multiview} (left), during inference, our architecture processes each view independently at the mask-decoding stage, while the visual geometry model still receives all views jointly.
Upon analyzing the behavior of MV-SAM, we observed that \picube produces highly consistent predictions whether frames are processed individually or as a set.
This consistency allows us to approximate the multi-view prediction task using single-view predictions that process all viewpoints independently. 
Consequently, MV-SAM can be trained with only single-view supervision, yet still generalizes naturally to multi-view prediction at inference time thanks to the stable geometric outputs from \picube.
}

\section{Comparison across Different Visual Geometry Models}

\revision{
Throughout the MV-SAM experiments, we used $\pi^3$~\citep{wang2025pi3} as our default visual geometry model (VGM), as it provides strong geometric priors and preserves equivariance under frame permutations. 
To assess how different VGMs influence MV-SAM’s performance, we compared three VGMs--$\pi^3$, VGGT~\citep{wang2025vggt}, and WorldMirror~\citep{liu2025worldmirror}--under an identical training setup. 
We follow the evaluation in the main paper on DL3DV experiments.
As reported in Table~\ref{tab:vgms}, both $\pi^3$ and WorldMirror yield clear performance improvements over VGGT.
}
\revision{
As reported in WorldMirror, WorldMirror achieves reconstruction quality comparable to $\pi^3$ across various benchmarks while consistently outperforming VGGT. Our results follow the same tendency: MV-SAM exhibits consistent performance gains when paired with either $\pi^3$ or WorldMirror, compared to VGGT. These findings highlight that the capability of the underlying visual geometry model directly affects MV-SAM’s final prediction performance. 
Therefore, MV-SAM has the potential to be further enhanced with the development of stronger visual geometry models.
}

\begin{table}[!ht]
\centering
\caption{\revision{Performance change of using different VGMs}}
\vspace{1mm}
\begin{tabular}{lcc}
\toprule
 & mIoU (\%) & mAcc (\%) \\
\midrule
VGGT ~\citep{wang2025vggt} & 61.1 & 90.4 \\
WorldMirror ~\citep{liu2025worldmirror} & 74.3 & 92.6 \\
$\pi^3$ ~\citep{wang2025pi3} & \textbf{75.1} & \textbf{91.8} \\
\bottomrule
\end{tabular}
\label{tab:vgms}
\end{table}

\revision{
We conduct a control experiment to evaluate the robustness of MV-SAM against noise in the reconstructed pointmaps generated by visual geometry models. Specifically, we perturb the pointmaps by adding Gaussian noise of varying magnitudes and assess the resulting performance degradation on ScanNet++. The noise scale is defined relative to the standard deviation of the pointmap coordinates; a scale of 1.0 corresponds to adding Gaussian noise with standard deviation to each coordinate.
}

\revision{
As shown in Table~\ref{tab:noises}, MV-SAM remains stable under moderate noise levels and maintains strong performance up to a noise scale of 0.5. Interestingly, even under extremely large perturbations--such as a noise scale of 4.0--the model still retains the ability to detect objects by leveraging the strong image embeddings provided by the pre-trained SAM2 encoder. Nevertheless, as expected, the overall performance gradually degrades as the noise magnitude increases.\
}

\begin{table}[!ht]
\centering
\caption{\revision{Performance change when injecting noises on predicted pointmaps. }}
\vspace{1mm}
\begin{tabular}{ccc}
\toprule
noise scale $\sigma$ & mIoU (\%) & mAcc (\%) \\
\midrule
0.0 & 48.9 & 63.5 \\
0.25 & 48.4 & 64.2 \\
0.5 & 47.1 & 65.1 \\
1.0 & 41.5 & 59.2 \\
2.0 & 39.7 & 58.4 \\
4.0 & 33.1 & 49.2 \\
\bottomrule
\end{tabular}
\label{tab:noises}
\end{table}
\section{Model Statistics}
\revision{
We report the statistics for running MV-SAM and compare them against various baselines--SAGA~\citep{SAGA}, OmniSeg3D~\citep{OmniSeg3D}, and SAM2-Video~\citep{SAM2}. For a fair comparison, we use 20 frames and measure the pre-processing time for each scene and the inference time for each query. In detail, the pre-processing time for per-scene optimization corresponds to the per-scene training time, whereas for training-free approaches such as SAM2-Video and MV-SAM, it corresponds to their feed-forward computation. As shown in Table~\ref{tab:spec_time_compare}, per-scene optimization approaches require substantial training time to preprocess features for each scene, while training-free approaches enable fast pre-processing due to their feed-forward predictions.
}

\revision{
Nevertheless, per-scene optimization approaches achieve faster inference, as they pre-extract features specific to the scene, whereas training-free approaches require longer inference time. We also observe that our model achieves faster inference than SAM2-Video, since MV-SAM processes all views simultaneously, while SAM2-Video requires iterative mask predictions, which incur linear complexity with respect to the number of frames.
}

\begin{table}[!ht]
\centering
\caption{\revision{Training and inference time on a DL3DV scene with 20 frames.}}
\vspace{1mm}
\begin{tabular}{ccc}
\toprule
 & Pre-processing & Inference \\
\midrule
SAGA~\citep{SAGA} & 31 (min) & 528 (ms) \\
OmniSeg3D~\citep{OmniSeg3D} & 37 (min) & 463 (ms) \\
SAM2-Video~\citep{SAM2} & 3.2 (s) & 4.8 (s) \\
MV-SAM & 5.1 (s) & 1.1 (s) \\
\bottomrule
\end{tabular}
\label{tab:spec_time_compare}
\end{table}

\revision{
Additionally, we compare the number of `trainable' parameters and required FLOPs between SAM2-Video and MV-SAM. Since both MV-SAM and SAM2-Video freeze their encoders during training, we only compare the number of learnable parameters. 
We use the same setup as in Table~\ref{tab:spec_time_compare}. 
As shown in Table~\ref{tab:spec_compute_compare}, MV-SAM requires more FLOPs due to the heavy computation of visual geometry models (VGMs). However, MV-SAM does not have the memory modules proposed in SAM2-Video, so ours use less number of trainable parameters than those in SAM2-Video.
% by pre-computing 3D positions for all frames, our model achieves faster inference than SAM2-Video, which is more beneficial in user scenarios where multiple clicks are sequentially provided by the user.
}

\begin{table}[!ht]
\centering
\caption{\revision{The number of parameters and FLOPs of SAM2-Video and MV-SAM}}
\vspace{1mm}
\begin{tabular}{ccc}
\toprule
 & \# parameters & FLOPs (TFLOPs) \\
\midrule
SAM2-Video~\citep{SAM2} & 12.3M & 16.8 \\
MV-SAM (ours) & 4.1M & 44.6 \\
\bottomrule
\end{tabular}
\label{tab:spec_compute_compare}
\end{table}

\section{Additional Baselines}

\revision{
We additionally compare more recent baselines, SAM2-Long~\citep{ding2024sam2long} and SAM3~\citep{carion2025sam3segmentconcepts}, with MV-SAM with the identical setup in Table~\ref{tab:sam2-ours-prompts}. 
Following our evaluation protocol, we benchmarked both methods on three datasets: ScanNet++, uCo3D, and DL3DV. 
Additionally, we included a simple baseline that unprojects user prompts into 3D and re-projects them into each image to run SAM2 image predictor~\citep{SAM2} independently per view. 
}

\begin{table}[t]
\centering
\caption{Comparison of mIoU and mAcc on the ScanNet++, uCo3D, and DL3DV benchmarks. Note that Prompt Projection$^\dagger$ is a simple baseline that unprojects user prompts into 3D and re-projects those prompts into each image to run the SAM2 image predictor independently per view. }
\label{tab:benchmark_results}
\resizebox{1.0\linewidth}{!}{
\begin{tabular}{lcccccc}
\toprule
\multirow{2}{*}{Method} &
\multicolumn{2}{c}{ScanNet++} &
\multicolumn{2}{c}{uCo3D} &
\multicolumn{2}{c}{DL3DV} \\
\cmidrule(lr){2-3}
\cmidrule(lr){4-5}
\cmidrule(lr){6-7}
 & mIoU ($\uparrow$) & mAcc ($\uparrow$) & mIoU ($\uparrow$) & mAcc ($\uparrow$) & mIoU ($\uparrow$) & mAcc ($\uparrow$) \\
\midrule
Prompt Projection$^\dagger$ & 0.292 & 0.592 & 0.782 & 0.833 & 0.412 & 0.702 \\
SAM2-Video~\citep{SAM2} & 0.461 & 0.614 & 0.819 & 0.913 & 0.673 & 0.829 \\
SAM2-Long~\citep{ding2024sam2long} & 0.415 & 0.614 & 0.729 & 0.864 & 0.605 & 0.785 \\
SAM3~\citep{carion2025sam3segmentconcepts} & 0.486 & 0.634 & 0.824 & 0.914 & 0.681 & 0.826 \\
\midrule
\textbf{MV-SAM (Ours)} & \textbf{0.489} & \textbf{0.635} & \textbf{0.877} & \textbf{0.950} & \textbf{0.751} & \textbf{0.918} \\
\bottomrule
\end{tabular}
}
\end{table}

\revision{
Across all benchmarks, our MV-SAM consistently outperforms the aforementioned baselines, even though these methods are trained on large-scale annotated video datasets. 
This demonstrates the benefit of explicitly incorporating 3D awareness into the model. 
Moreover, naively projecting prompts into all views introduces significant challenges in occluded scenes; on ScanNet++, in particular, we observe substantial performance degradation due to the frequent presence of occlusions. 
Finally, we note that even the latest approach, SAM3, still exhibits inconsistent tracking, despite being trained on complex datasets designed to enhance SAM2.
}
\section{Additional Experiments}
In this section, we list additional experiments to check our model's performance. 
We follow the same setup with the control experiments in our main paper. 

\paragraph{Loss for training}
In Table ~\ref{tab:loss_comparison}, we compare our model using different loss functions, including ASL~\citep{benbaruch2020asymmetric}, binary cross entropy (BCE), and focal loss~\citep{lin2017focal}, both with and without the addition of Dice loss. To ensure fairness, all hyperparameters are kept identical across loss configurations.

\begin{table}[!ht]
\centering
\caption{Comparison of different loss functions.}
\vspace{1mm}
\begin{tabular}{lcc}
\toprule
Loss type & mIoU (\%) ($\uparrow$) & mAcc (\%) ($\uparrow$) \\
\midrule
ASL~\citep{benbaruch2020asymmetric}                 & 43.3 & \textbf{67.7} \\
ASL~\citep{benbaruch2020asymmetric} + Dice       & 39.4 & 67.2 \\
BCE                 & 44.0 & 62.4 \\
BCE + Dice          & 49.3 & 64.5 \\
Focal~\citep{lin2017focal}               & 47.1 & 66.1 \\
Focal~\citep{lin2017focal} + Dice (Ours) & 
\textbf{52.2} & 66.7 \\
\bottomrule
\end{tabular}
\label{tab:loss_comparison}
\end{table}

\paragraph{3D Decoder}

Another approach is to employ 3D networks as the mask decoder.
This design introduces a 3D inductive bias, encouraging neighboring points to produce similar logits.
We observed that the grid size strongly affects the final performance, but due to memory constraints, the finest feasible resolution was limited to 0.005.
Consistent with the observation in the main paper, voxelization enforces 3D consistency; however, the lack of metric alignment prevents the model from generalizing well across diverse scenes as shown in Table~\ref{tab:mask_decoder}.

\begin{table}[!h]
\centering
\caption{Ablation results with different 3D networks and grid resolutions.}
\begin{tabular}{lccc}
\toprule
Mask Decoder & Grid & mIoU ($\uparrow$) & mAcc ($\uparrow$) \\
\midrule
             & 0.05  & 44.4 & 61.4 \\
Minkowski    & 0.01  & 46.1 & 60.2 \\
             & 0.005 & 46.2 & 60.3 \\
\midrule
             & 0.05  & 44.1 & 62.2 \\
PTv3         & 0.01  & 44.2 & 61.0 \\
             & 0.005 & 45.4 & 62.4 \\
\midrule
Ours         &   -   & \textbf{52.2} & \textbf{66.7} \\
\bottomrule
\end{tabular}
\label{tab:mask_decoder}
\end{table}

\paragraph{Different numbers of frames.}
As discussed in the main paper, our single-view attention outperforms full-view attention, which involves all viewpoints during the attention operations. 
Below, we provide the exact numerical values corresponding to the graph in Figure~\ref{fig:exp-num-frames} in Table~\ref{table:frame_changes}.

\begin{table}[!h]
\vspace{2mm}
\caption{Performance comparison across different numbers of frames.}
\centering
\begin{tabular}{lccccccc}
\toprule
\# frames & 2 & 4 & 8 & 16 & 32 & 100 \\
\midrule
Single-view (Ours) & 54.3  & 55.2 & 53.1 & 53.2 & 52.1 & 52.2 \\
Full-view   & 54.8  & 55.0 & 54.5 & 51.6 & 47.6 & 45.8 \\
\bottomrule
\end{tabular}
\label{table:frame_changes}
\end{table}

\paragraph{Confidence-aware Prompting}

We further investigate the proportion of points that should be treated as low-confidence. By varying the low-confidence thresholds, we compare mIoU and mAcc on ScanNet++. We observe that discarding more than \revision{15}\% of the points leads to performance degradation, indicating that removing too many reliable points can negatively impact the results as shown in Table~\ref{tab:CGPE_supp}.

\begin{table}[!h]
\caption{Control experiments by varying the confidence threshold.}
\centering
\begin{tabular}{ccc}
\toprule
Confidence Threshold & mIoU (\%) ($\uparrow$) & mAcc (\%) ($\uparrow$) \\
\midrule
0    & 44.45 & 61.11 \\
0.05 & 51.65 & 66.53 \\
0.10 & 52.18 & \textbf{66.81} \\
0.15 & \textbf{52.25} & 66.70 \\
0.20 & 50.43 & 66.45 \\
0.25 & 49.55 & 66.18 \\
0.30 & 49.82 & 66.46 \\
\bottomrule
\end{tabular}
\label{tab:CGPE_supp}
\end{table}

\paragraph{Effect of standardization}
\label{subsec:standard}

We apply standardization before extracting sinusoidal embeddings. 
Experiments show that this strategy improves robustness to different scenarios, such as an increasing number of frames. 
As shown in Table~\ref{tab:standard}, our model consistently outperforms its variant without standardization. 
Without standardization, we observe many failure cases, particularly in outdoor scenes. 
We conjecture that this is because outdoor scenes contain more widely distributed points, which makes the model more prone to inferring outliers during evaluation.

\begin{table}[!h]
\caption{Effect of using standardization.}
\centering
\begin{tabular}{ccc}
\toprule
Standardization & w/ (Ours)  & w/o  \\
\midrule
ScanNet++ & \textbf{49.8} / \textbf{63.6} & 27.3 / 51.2 \\
uCo3D & \textbf{86.9} / \textbf{94.4} & 16.2 / 24.5 \\
DL3DV & \textbf{35.7} / \textbf{64.7} & 6.2 / 13.3 \\
\bottomrule
\end{tabular}
\label{tab:standard}
\end{table}

% \clearpage

\section{Per-scene Results for NVOS and SPIn-NeRF}
\label{subsec:perscene_results}
Since we excluded the \textit{orchid} scene from NVOS and the \textit{pinecone} scene from SPIn-NeRF, 
we provide per-scene results for SAM2, MV-SAM, and prior baselines~\citep{spinnerf, SA3D, SA3D-GS, SAGA, OmniSeg3D} to facilitate future research. 
Per-scene mIoU and mAcc are reported in Table~\ref{tab:per_scene_nvos} for NVOS and in Table~\ref{tab:per_scene_spinnerf}.

\begin{table}[!h]
\centering
\resizebox{\linewidth}{!}{
\begin{tabular}{lcccccccc}
\toprule
Method & fern & flower & fortress & horns\_center & horns\_left & leaves & trex & Avg. \\
\midrule
\multirow{2}{*}{SA3D~\citep{SA3D}} 
 & 82.90 & 94.60 & 98.30 & 96.20 & 90.20 & 93.20 & 81.99 & 91.06 \\
 & 94.39 & 98.74 & 99.68 & 99.33 & 99.36 & 99.57 & 97.41 & 98.35 \\
\midrule
\multirow{2}{*}{SA3D-GS~\citep{SA3D-GS}} 
 & 85.50 & 96.79 & 98.07 & 98.18 & 94.33 & 93.67 & 82.13 & 92.67 \\
 & 95.30 & 99.21 & 99.64 & 99.52 & 99.55 & 99.60 & 97.47 & 98.51 \\
\midrule
\multirow{2}{*}{SAGA~\citep{SAGA}} 
 & 83.53 & 96.62 & 98.16 & 98.06 & 93.59 & 93.51 & 80.81 & 92.57 \\
 & 94.60 & 99.17 & 99.65 & 99.50 & 99.51 & 99.59 & 97.26 & 98.55 \\
\midrule
\multirow{2}{*}{OmniSeg3D~\citep{OmniSeg3D}} 
 & 82.70 & 95.30 & 98.50 & 97.70 & 95.60 & 92.70 & 87.40 & 92.84 \\
 & 94.30 & 98.90 & 99.70 & 99.60 & 99.70 & 99.50 & 98.30 & 98.57 \\
\midrule
\multirow{2}{*}{SAM2~\citep{SAM2}} 
 & 82.83 & 95.20 & 97.03 & 95.71 & 94.66 & 93.37 & 62.24 & 88.72 \\
 & 93.94 & 97.75 & 98.52 & 97.93 & 95.80 & 96.84 & 81.66 & 94.63 \\
\midrule
\multirow{2}{*}{MV-SAM} 
 & 82.90 & 95.50 & 97.50 & 97.60 & 94.50 & 94.30 & 82.50 & 92.11 \\
 & 94.90 & 98.30 & 98.70 & 97.50 & 98.40 & 98.50 & 95.90 & 97.46 \\
\bottomrule
\end{tabular}
}
\caption{Per-scene quantitative results on NVOS. The first row of each method corresponds to metric mIoU and the second row to metric mAcc.}
\label{tab:per_scene_nvos}
\end{table}

\begin{table}[!h]
\centering
\resizebox{\linewidth}{!}{
\begin{tabular}{lccccccc}
\toprule
Scene & SPIn-NeRF & SA3D & SA3D-GS & SAGA & OmniSeg3D & SAM-Video & Ours \\
\midrule
room     & 95.6 / 99.4 & 88.22 / 98.33 & 93.73 / 99.18 & 96.91 / 99.59 & 97.9 / 99.7 & 91.90 / 96.00 & 91.50 / 95.90 \\
orchids  & 92.7 / 98.8 & 83.55 / 96.86 & 84.68 / 97.18 & 90.55 / 98.29 & 92.3 / 98.7 & 86.90 / 94.32 & 83.50 / 95.70 \\
horns    & 92.8 / 98.7 & 94.49 / 99.02 & 95.26 / 99.17 & 92.96 / 98.71 & 91.5 / 98.5 & 84.44 / 92.39 & 89.60 / 95.20 \\
fern     & 94.3 / 99.2 & 97.05 / 99.59 & 96.67 / 99.54 & 96.49 / 99.51 & 97.5 / 99.7 & 97.33 / 98.88 & 97.40 / 99.00 \\
fortress & 97.7 / 99.7 & 98.33 / 99.75 & 98.06 / 99.71 & 96.16 / 99.41 & 97.9 / 99.7 & 69.49 / 84.76 & 98.10 / 99.10 \\
leaves   & 94.9 / 99.7 & 97.18 / 99.85 & 97.20 / 99.85 & 95.52 / 99.75 & 96.0 / 99.8 & 96.76 / 98.53 & 95.80 / 98.10 \\
fork     & 87.9 / 99.5 & 89.41 / 99.55 & 87.91 / 99.49 & 85.84 / 99.42 & 90.4 / 99.6 & 83.88 / 92.67 & 90.00 / 96.00 \\
truck    & 85.2 / 95.1 & 90.82 / 96.66 & 94.80 / 98.20 & 95.71 / 98.53 & 96.1 / 98.7 & 83.49 / 91.86 & 96.40 / 98.60 \\
lego     & 74.9 / 99.2 & 92.15 / 99.75 & 91.89 / 99.75 & 93.17 / 99.79 & 90.8 / 99.7 & 85.25 / 93.01 & 89.90 / 96.90 \\
\midrule
Avg.     & 90.67 / 98.81 & 92.36 / 98.82 & 93.36 / 99.12 & 93.70 / 99.22 & 94.49 / 99.34 & 86.60 / 93.60 & 92.47 / 97.17 \\
\bottomrule
\end{tabular}}
\caption{Per-scene comparison of mIoU / mAcc across different methods on the SPIn-NeRF dataset. Average (Avg.) values are shown in the last row.}
\label{tab:per_scene_spinnerf}
\end{table}

% \clearpage

\section{Qualitative Results}

\begin{figure}[!t]
    \centering
    \includegraphics[width=\linewidth]{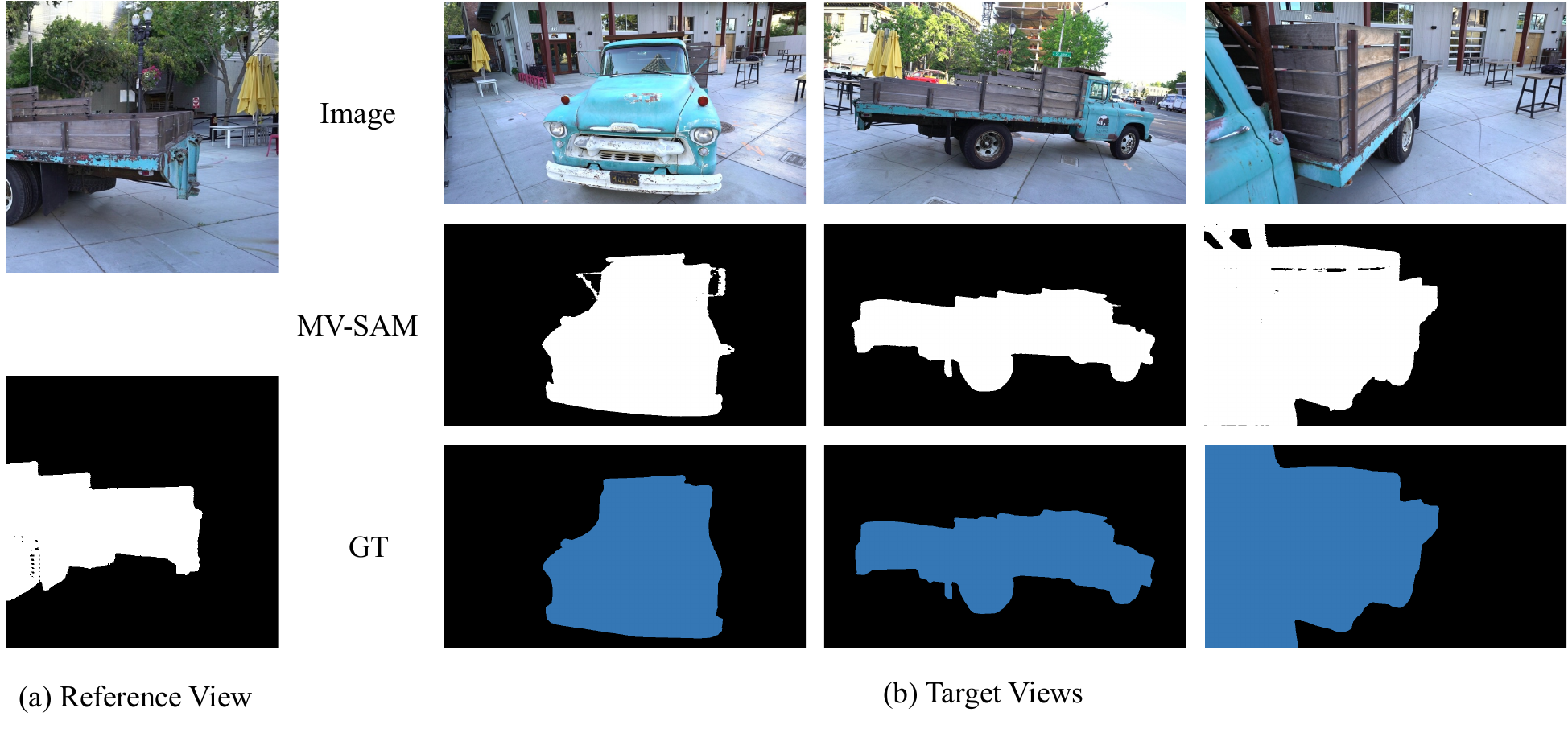}
    \vspace{-6mm}
    \caption{Visualization of predicted results of MV-SAM for partially occluded cases. We explicitly crop the reference view to encorce scenarios in which the object is only partially observed. MV-SAM reliably reconstructs the complete object masks despite the incomplete reference cues due to the 3D awareness. }
    \label{fig:supp_occ}
\end{figure}

\begin{figure*}[!t]
    \centering
    \includegraphics[width=\linewidth]{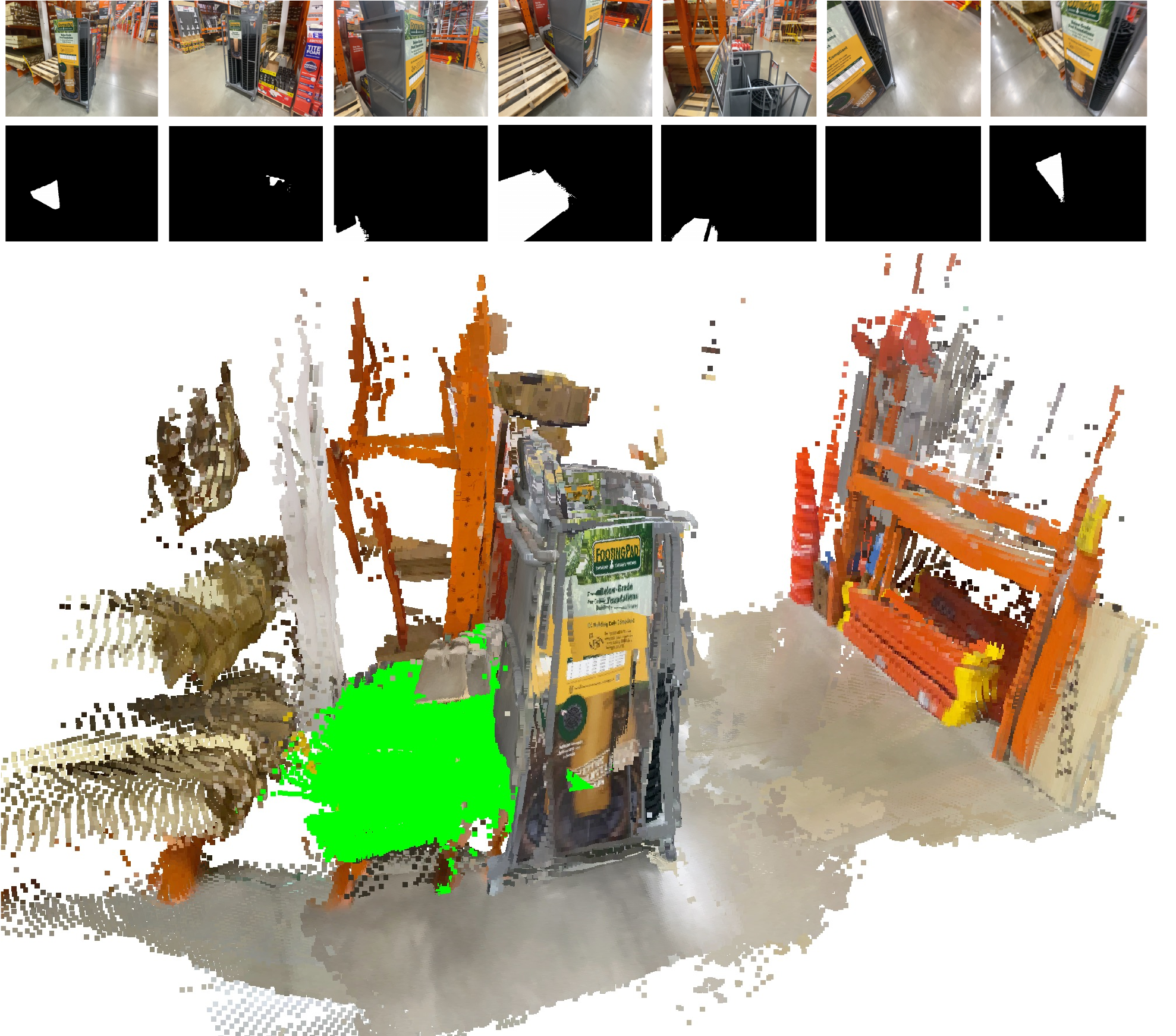}
    \vspace{-6mm}
    \caption{Magnified qualitative results on a DL3DV scene. Owing to the large scale of the scene, detailed predictions are difficult to show in the main paper, so we present magnified results here. The highlighted region in green demonstrates that our method correctly identifies the wood block in the scene. }
    \label{fig:supp_large_scale}
    \vspace{-4mm}
\end{figure*}

\begin{figure}[!t]
    \centering
    \includegraphics[width=\linewidth]{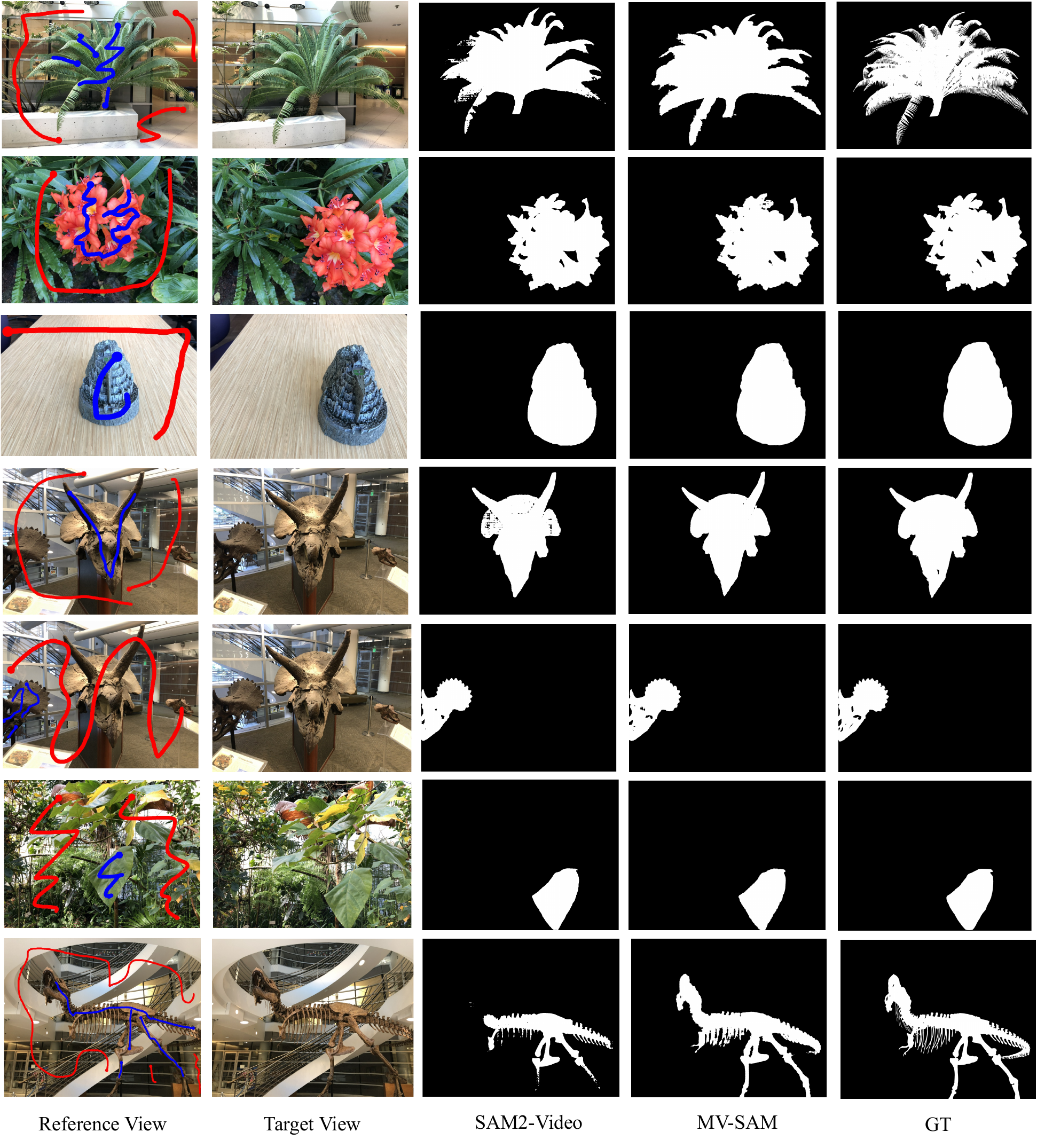}
    \caption{Visualization of predicted masks of SAM2-Video and MV-SAM from the NVOS dataset. }
    \label{fig:supp_nvos}
\end{figure}

\begin{figure}[!t]
    \centering
    \includegraphics[width=\linewidth]{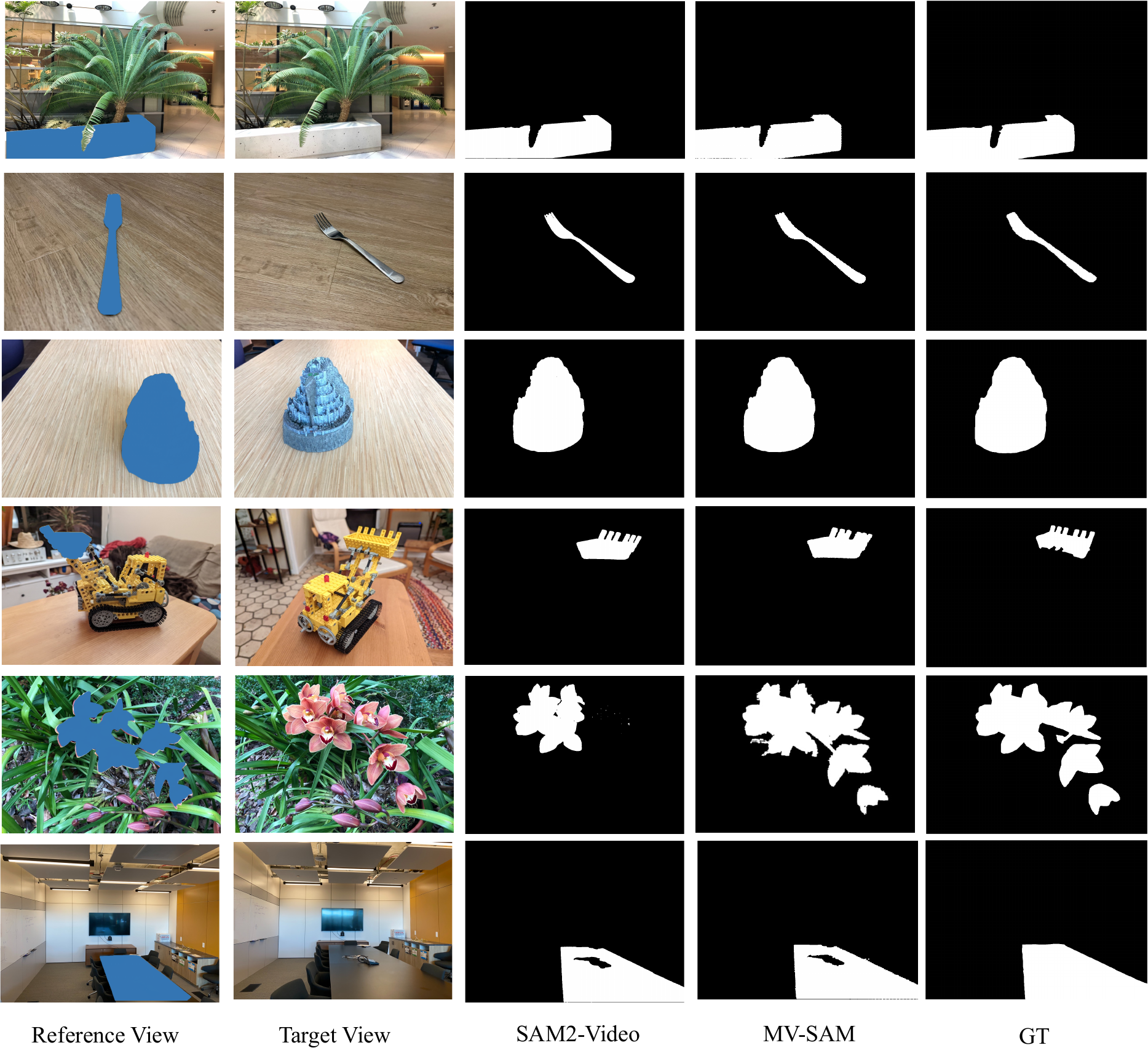}
    \caption{Visualization of predicted masks of SAM2-Video and MV-SAM from the SPIn-NeRF dataset. }
    \label{fig:supp_mvseg}
\end{figure}

\begin{figure}[!t]
    \centering
    \includegraphics[width=\linewidth]{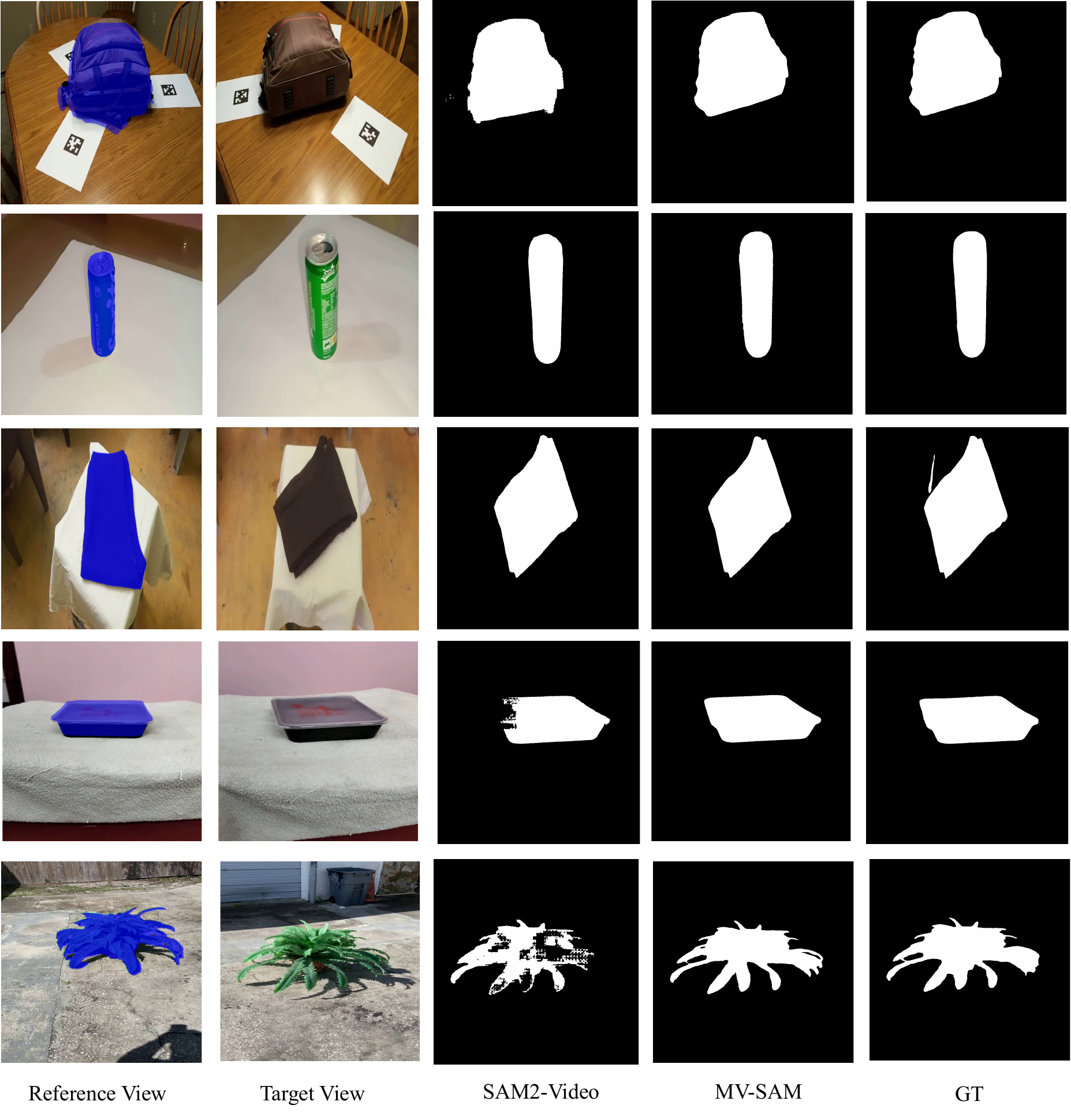}
    \caption{Visualization of predicted masks of SAM2-Video and MV-SAM from the uCo3D dataset. }
    \label{fig:supp_uco3d}
\end{figure}

In this section, we present additional qualitative results.
\revision{
We provide an additional toy experiment to assess whether MV-SAM can recover whole objects when the reference image contains only partial observations.
For clearer analysis, we explicitly crop the reference view to enforce scenarios in which the object is only partially visible.
We then visualize the corresponding predictions on multiple target views captured from diverse viewpoints.
As shown in Figure~\ref{fig:supp_occ}, MV-SAM reliably reconstructs complete object masks even under incomplete reference cues, facilitated by its 3D-aware model design.
}
Figure~\ref{fig:supp_large_scale} shows scene-level pointmaps and their corresponding predicted masks, highlighting the effectiveness of MV-SAM in large-scale sequences.
Figures~\ref{fig:supp_nvos}, \ref{fig:supp_mvseg}, and \ref{fig:supp_uco3d} provide further visualizations on NVOS, MVSeg, and uCo3D datasets, respectively.
\revision{
Lastly, we include several video results in the supplementary materials, where MV-SAM consistently demonstrates superior performance compared to SAM2-Video.
} 

% \clearpage

\section{Limitation}
\revision{
Our model also encounters difficulties in more challenging scenarios, such as scenes containing a large number of dynamic objects that break the assumption of static geometry.
Similarly, in non-3D-based domains like cartoon animations or synthetic imagery without a consistent underlying geometry, the model struggles to maintain accurate segmentation.
}

\begin{figure}[!h]
    \centering
    \includegraphics[width=1.0\linewidth]{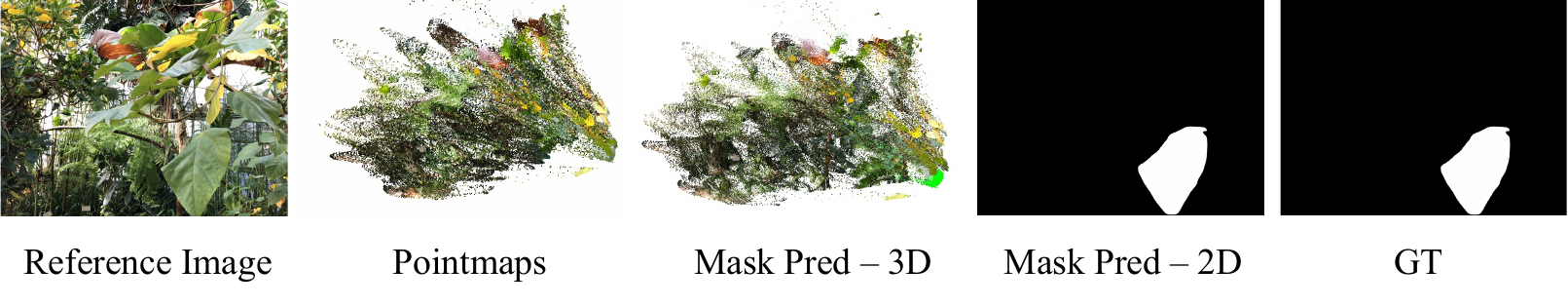}
    \caption{Failure cases, where the predicted 2D masks remain clear, but the 3D geometry estimated by \picube is inaccurate. }
    \label{fig:failure}
\end{figure}

\revision{
Nevertheless, we observe that our approach can still produce high-quality masks even when the pointmaps are imperfect, as illustrated in Figure~\ref{fig:failure}. 
For instance, in ScanNet++ scenes with reflective surfaces or partial occlusions, the resulting pointmaps may contain substantial noise. 
Despite this, our model reliably reconstructs object masks that remain well aligned with the underlying scene structure. 
This demonstrates a notable degree of robustness to imperfect geometric priors. 
Further enhancing this robustness--either by incorporating strategies that explicitly model uncertainty or by leveraging improved pointmap estimation methods--represents a promising future work direction.
}

\label{sec:conclusion}

% \textbf{Links:} \hspace{2pt}
% {
% \hypersetup{urlcolor=nvidiagreen}
% \href{https://nvlabs.github.io/EAGLE/}{Project Page}
% }

\clearpage
\setcitestyle{numbers}
\bibliographystyle{plainnat}
\bibliography{iclr2026_conference}

\begin{thebibliography}{47}
\providecommand{\natexlab}[1]{#1}
\providecommand{\url}[1]{\texttt{#1}}
\expandafter\ifx\csname urlstyle\endcsname\relax
  \providecommand{\doi}[1]{doi: #1}\else
  \providecommand{\doi}{doi: \begingroup \urlstyle{rm}\Url}\fi

\bibitem[Bekuzarov et~al.(2023)Bekuzarov, Bermudez, Lee, and Li]{bekuzarov2023xmempp}
Maksym Bekuzarov, Ariana Bermudez, Joon-Young Lee, and Hao Li.
\newblock Xmem++: Production-level video segmentation from few annotated frames.
\newblock In \emph{Proceedings of the IEEE/CVF International Conference on Computer Vision (ICCV)}, 2023.

\bibitem[Ben‐Baruch et~al.(2020)Ben‐Baruch, Ridnik, Zamir, Noy, Friedman, Protter, and Zelnik-Manor]{benbaruch2020asymmetric}
Emanuel Ben‐Baruch, Tal Ridnik, Nadav Zamir, Asaf Noy, Itamar Friedman, Matan Protter, and Lihi Zelnik-Manor.
\newblock Asymmetric loss for multi‐label classification, 2020.

\bibitem[Carion et~al.(2025)Carion, Gustafson, Hu, Debnath, Hu, Suris, Ryali, Alwala, Khedr, Huang, Lei, Ma, Guo, Kalla, Marks, Greer, Wang, Sun, Rädle, Afouras, Mavroudi, Xu, Wu, Zhou, Momeni, Hazra, Ding, Vaze, Porcher, Li, Li, Kamath, Cheng, Dollár, Ravi, Saenko, Zhang, and Feichtenhofer]{carion2025sam3segmentconcepts}
Nicolas Carion, Laura Gustafson, Yuan-Ting Hu, Shoubhik Debnath, Ronghang Hu, Didac Suris, Chaitanya Ryali, Kalyan~Vasudev Alwala, Haitham Khedr, Andrew Huang, Jie Lei, Tengyu Ma, Baishan Guo, Arpit Kalla, Markus Marks, Joseph Greer, Meng Wang, Peize Sun, Roman Rädle, Triantafyllos Afouras, Effrosyni Mavroudi, Katherine Xu, Tsung-Han Wu, Yu~Zhou, Liliane Momeni, Rishi Hazra, Shuangrui Ding, Sagar Vaze, Francois Porcher, Feng Li, Siyuan Li, Aishwarya Kamath, Ho~Kei Cheng, Piotr Dollár, Nikhila Ravi, Kate Saenko, Pengchuan Zhang, and Christoph Feichtenhofer.
\newblock Sam 3: Segment anything with concepts, 2025.
\newblock URL \url{https://arxiv.org/abs/2511.16719}.

\bibitem[Cen et~al.(2023{\natexlab{a}})Cen, Fang, Yang, Xie, Zhang, Shen, and Tian]{SAGA}
Jiazhong Cen, Jiemin Fang, Chen Yang, Lingxi Xie, Xiaopeng Zhang, Wei Shen, and Qi~Tian.
\newblock Segment any 3d gaussians.
\newblock \emph{arXiv preprint}, 2023{\natexlab{a}}.
\newblock arXiv:2312.00860.

\bibitem[Cen et~al.(2023{\natexlab{b}})Cen, Zhou, Fang, Yang, Shen, Xie, Zhang, and Tian]{SA3D}
Jiazhong Cen, Zanwei Zhou, Jiemin Fang, Chen Yang, Wei Shen, Lingxi Xie, Xiaopeng Zhang, and Qi~Tian.
\newblock Segment anything in 3d with nerfs.
\newblock In \emph{Advances in Neural Information Processing Systems (NeurIPS)}, 2023{\natexlab{b}}.

\bibitem[Cen et~al.(2025)Cen, Fang, Zhou, Yang, Xie, Zhang, Shen, and Tian]{SA3D-GS}
Jiazhong Cen, Jiemin Fang, Zanwei Zhou, Chen Yang, Lingxi Xie, Xiaopeng Zhang, Wei Shen, and Qi~Tian.
\newblock Segment anything in 3d with radiance fields (sa3d): Adaptation to 3d-gaussian splatting for real-time segmentation.
\newblock \emph{International Journal of Computer Vision}, 2025.

\bibitem[Cheng and Schwing(2022)]{cheng2022xmem}
Ho~Kei Cheng and Alexander~G. Schwing.
\newblock Xmem: Long-term video object segmentation with an atkinson–shiffrin memory model.
\newblock In \emph{European Conference on Computer Vision (ECCV)}, pages 640--658, 2022.
\newblock \doi{10.1007/978-3-031-19815-1\_37}.

\bibitem[Cheng et~al.(2023{\natexlab{a}})Cheng, Oh, Price, Schwing, and Lee]{cheng2023tracking}
Ho~Kei Cheng, Seoung~Wug Oh, Brian~L. Price, Alexander Schwing, and Joon-Young Lee.
\newblock Tracking anything with decoupled video segmentation.
\newblock In \emph{Proceedings of the IEEE/CVF International Conference on Computer Vision (ICCV)}, pages 1316--1326, 2023{\natexlab{a}}.

\bibitem[Cheng et~al.(2024)Cheng, Oh, Price, Lee, and Schwing]{cheng2023putting}
Ho~Kei Cheng, Seoung~Wug Oh, Brian Price, Joon-Young Lee, and Alexander Schwing.
\newblock Putting the object back into video object segmentation (cutie).
\newblock In \emph{Proceedings of the IEEE/CVF Conference on Computer Vision and Pattern Recognition (CVPR)}, 2024.

\bibitem[Cheng et~al.(2023{\natexlab{b}})Cheng, Li, Xu, Li, Yang, Wang, and Yang]{cheng2023segment}
Yangming Cheng, Liulei Li, Yuanyou Xu, Xiaodi Li, Zongxin Yang, Wenguan Wang, and Yi~Yang.
\newblock Segment and track anything.
\newblock \emph{arXiv preprint arXiv:2305.06558}, 2023{\natexlab{b}}.

\bibitem[Choi et~al.(2025)Choi, Song, Kim, Kim, and Do]{Click-Gaussian}
Seokhun Choi, Hyeonseop Song, Jaechul Kim, Taehyeong Kim, and Hoseok Do.
\newblock Click-gaussian: Interactive segmentation to any 3d gaussians.
\newblock In \emph{European Conference on Computer Vision (ECCV)}, pages 289--305, 2025.

\bibitem[Choy et~al.(2019)Choy, Gwak, and Savarese]{choy2019minkowski}
Christopher~B. Choy, JunYoung Gwak, and Silvio Savarese.
\newblock 4d spatio-temporal convnets: Minkowski convolutional neural networks.
\newblock In \emph{Proceedings of the IEEE/CVF Conference on Computer Vision and Pattern Recognition (CVPR)}, pages 3075--3084, 2019.

\bibitem[Ding et~al.(2024)Ding, Qian, Dong, Zhang, Zang, Cao, Guo, Lin, and Wang]{ding2024sam2long}
Shuangrui Ding, Rui Qian, Xiaoyi Dong, Pan Zhang, Yuhang Zang, Yuhang Cao, Yuwei Guo, Dahua Lin, and Jiaqi Wang.
\newblock Sam2long: Enhancing sam 2 for long video segmentation with a training-free memory tree.
\newblock \emph{arXiv preprint arXiv:2410.16268}, 2024.

\bibitem[Kerbl et~al.(2023)Kerbl, Kopanas, Leimk{\"u}hler, and Drettakis]{3dgs}
Bernhard Kerbl, Georgios Kopanas, Thomas Leimk{\"u}hler, and George Drettakis.
\newblock 3d gaussian splatting for real-time radiance field rendering.
\newblock \emph{ACM Trans. Graph.}, 42\penalty0 (4):\penalty0 139--1, 2023.

\bibitem[Kirillov et~al.(2023)Kirillov, Mintun, Ravi, Mao, Rolland, Gustafson, Xiao, Whitehead, Berg, Lo, et~al.]{SAM}
Alexander Kirillov, Eric Mintun, Nikhila Ravi, Hanzi Mao, Chloe Rolland, Laura Gustafson, Tete Xiao, Spencer Whitehead, Alexander~C Berg, Wan-Yen Lo, et~al.
\newblock Segment anything.
\newblock In \emph{Proceedings of the IEEE/CVF international conference on computer vision}, pages 4015--4026, 2023.

\bibitem[Leroy et~al.(2024)Leroy, Cabon, and Revaud]{leroy2024mast3r}
Vincent Leroy, Yohann Cabon, and Jérôme Revaud.
\newblock Grounding image matching in 3d with mast3r.
\newblock In \emph{European Conference on Computer Vision (ECCV)}, 2024.

\bibitem[Lin et~al.(2017)Lin, Goyal, Girshick, He, and Doll{\'a}r]{lin2017focal}
Tsung-Yi Lin, Priya Goyal, Ross Girshick, Kaiming He, and Piotr Doll{\'a}r.
\newblock Focal loss for dense object detection.
\newblock In \emph{Proceedings of the IEEE International Conference on Computer Vision (ICCV)}, pages 2980--2988, 2017.

\bibitem[Ling et~al.(2024)Ling, Sheng, Tu, Zhao, Xin, Wan, Yu, Guo, Yu, Lu, Li, Sun, Ashok, Mukherjee, Kang, Kong, Hua, Zhang, Benes, and Bera]{ling2024dl3dv10k}
Lu~Ling, Yichen Sheng, Zhi Tu, Wentian Zhao, Cheng Xin, Kun Wan, Lantao Yu, Qianyu Guo, Zixun Yu, Yawen Lu, Xuanmao Li, Xingpeng Sun, Rohan Ashok, Aniruddha Mukherjee, Hao Kang, Xiangrui Kong, Gang Hua, Tianyi Zhang, Bedrich Benes, and Aniket Bera.
\newblock Dl3dv-10k: A large-scale scene dataset for deep learning-based 3d vision.
\newblock In \emph{Proceedings of the IEEE/CVF Conference on Computer Vision and Pattern Recognition (CVPR)}, 2024.
\newblock URL \url{https://dl3dv-10k.github.io/DL3DV-10K/}.

\bibitem[Liu et~al.(2025{\natexlab{a}})Liu, Tayal, Wang, Zarzar, Monnier, Tertikas, Duan, Toisoul, Zhang, Neverova, Vedaldi, Shapovalov, and Novotny]{liu2025uco3d}
Xingchen Liu, Piyush Tayal, Jianyuan Wang, Jesus Zarzar, Tom Monnier, Konstantinos Tertikas, Jiali Duan, Antoine Toisoul, Jason~Y. Zhang, Natalia Neverova, Andrea Vedaldi, Roman Shapovalov, and David Novotny.
\newblock Uncommon objects in 3d.
\newblock \emph{arXiv preprint arXiv:2501.07574}, 2025{\natexlab{a}}.
\newblock URL \url{https://arxiv.org/abs/2501.07574}.

\bibitem[Liu et~al.(2025{\natexlab{b}})Liu, Min, Wang, Wu, Wang, Yuan, Luo, and Guo]{liu2025worldmirror}
Yifan Liu, Zhiyuan Min, Zhenwei Wang, Junta Wu, Tengfei Wang, Yixuan Yuan, Yawei Luo, and Chunchao Guo.
\newblock Worldmirror: Universal 3d world reconstruction with any-prior prompting.
\newblock \emph{arXiv preprint arXiv:2510.10726}, 2025{\natexlab{b}}.

\bibitem[Mildenhall et~al.(2021)Mildenhall, Srinivasan, Tancik, Barron, Ramamoorthi, and Ng]{nerf}
Ben Mildenhall, Pratul~P Srinivasan, Matthew Tancik, Jonathan~T Barron, Ravi Ramamoorthi, and Ren Ng.
\newblock Nerf: Representing scenes as neural radiance fields for view synthesis.
\newblock \emph{Communications of the ACM}, 65\penalty0 (1):\penalty0 99--106, 2021.

\bibitem[Milletari et~al.(2016)Milletari, Navab, and Ahmadi]{milletari2016vnet}
Fausto Milletari, Nassir Navab, and Seyed-Ahmad Ahmadi.
\newblock V-net: Fully convolutional neural networks for volumetric medical image segmentation.
\newblock \emph{Proceedings of 2016 Fourth International Conference on 3D Vision (3DV)}, pages 565--571, 2016.

\bibitem[Mirzaei et~al.(2023)Mirzaei, Aumentado-Armstrong, Derpanis, Kelly, Brubaker, Gilitschenski, and Levinshtein]{spinnerf}
Ashkan Mirzaei, Tristan Aumentado-Armstrong, Konstantinos~G. Derpanis, Jonathan Kelly, Marcus~A. Brubaker, Igor Gilitschenski, and Alex Levinshtein.
\newblock {SPIn-NeRF}: Multiview segmentation and perceptual inpainting with neural radiance fields.
\newblock In \emph{CVPR}, 2023.

\bibitem[Oquab et~al.(2023)Oquab, Darcet, Moutakanni, Vo, Szafraniec, Khalidov, Fernandez, Haziza, Massa, El-Nouby, et~al.]{oquab2023dinov2}
Maxime Oquab, Timoth{\'e}e Darcet, Th{\'e}o Moutakanni, Huy Vo, Marc Szafraniec, Vasil Khalidov, Pierre Fernandez, Daniel Haziza, Francisco Massa, Alaaeldin El-Nouby, et~al.
\newblock Dinov2: Learning robust visual features without supervision.
\newblock \emph{arXiv preprint arXiv:2304.07193}, 2023.

\bibitem[Pan et~al.(2024)Pan, Bar{\'a}th, Pollefeys, and Sch{\"o}nberger]{pan2024glomap}
Linfei Pan, D{\'a}niel Bar{\'a}th, Marc Pollefeys, and Johannes~Lutz Sch{\"o}nberger.
\newblock Global structure-from-motion revisited.
\newblock In \emph{European Conference on Computer Vision (ECCV)}, 2024.

\bibitem[Press et~al.(2022)Press, Smith, and Lewis]{press2024alibi}
Ofir Press, Noah~A Smith, and Mike Lewis.
\newblock Train short, test long: Attention with linear biases enables input length extrapolation.
\newblock In \emph{ICLR}, 2022.

\bibitem[Qi et~al.(2017{\natexlab{a}})Qi, Su, Mo, and Guibas]{pointnet}
Charles~R Qi, Hao Su, Kaichun Mo, and Leonidas~J Guibas.
\newblock Pointnet: Deep learning on point sets for 3d classification and segmentation.
\newblock In \emph{Proceedings of the IEEE conference on computer vision and pattern recognition}, pages 652--660, 2017{\natexlab{a}}.

\bibitem[Qi et~al.(2017{\natexlab{b}})Qi, Yi, Su, and Guibas]{qi2017pointnet++}
Charles~Ruizhongtai Qi, Li~Yi, Hao Su, and Leonidas~J Guibas.
\newblock Pointnet++: Deep hierarchical feature learning on point sets in a metric space.
\newblock \emph{Advances in neural information processing systems}, 30, 2017{\natexlab{b}}.

\bibitem[Raji{\v{c}} et~al.(2023)Raji{\v{c}}, Ke, Tai, Tang, Danelljan, and Yu]{rajic2023segment}
Frano Raji{\v{c}}, Lei Ke, Yu-Wing Tai, Chi-Keung Tang, Martin Danelljan, and Fisher Yu.
\newblock Segment anything meets point tracking.
\newblock \emph{arXiv preprint arXiv:2307.01197}, 2023.

\bibitem[Ravi et~al.(2024)Ravi, Gabeur, Hu, Hu, Ryali, Ma, Khedr, R{\"a}dle, Rolland, Gustafson, Mintun, Pan, Alwala, Carion, Wu, Girshick, Doll{\'a}r, and Feichtenhofer]{SAM2}
Nikhila Ravi, Valentin Gabeur, Yuan-Ting Hu, Ronghang Hu, Chaitanya Ryali, Tengyu Ma, Haitham Khedr, Roman R{\"a}dle, Chloe Rolland, Laura Gustafson, Eric Mintun, Junting Pan, Kalyan~Vasudev Alwala, Nicolas Carion, Chao-Yuan Wu, Ross Girshick, Piotr Doll{\'a}r, and Christoph Feichtenhofer.
\newblock Sam 2: Segment anything in images and videos.
\newblock \emph{arXiv preprint}, 2024.

\bibitem[Ren et~al.(2022)Ren, Agarwala, Russell, Schwing, and Wang]{NVOS}
Zhongzheng Ren, Aseem Agarwala, Bryan Russell, Alexander~G. Schwing, and Oliver Wang.
\newblock Neural volumetric object selection.
\newblock In \emph{Proceedings of the IEEE/CVF Conference on Computer Vision and Pattern Recognition (CVPR)}, pages 6133--6142, June 2022.

\bibitem[Sch\"{o}nberger and Frahm(2016)]{colmap1}
Johannes~Lutz Sch\"{o}nberger and Jan-Michael Frahm.
\newblock Structure-from-motion revisited.
\newblock In \emph{Conference on Computer Vision and Pattern Recognition (CVPR)}, 2016.

\bibitem[Sch\"{o}nberger et~al.(2016)Sch\"{o}nberger, Zheng, Pollefeys, and Frahm]{colmap2}
Johannes~Lutz Sch\"{o}nberger, Enliang Zheng, Marc Pollefeys, and Jan-Michael Frahm.
\newblock Pixelwise view selection for unstructured multi-view stereo.
\newblock In \emph{European Conference on Computer Vision (ECCV)}, 2016.

\bibitem[Sun et~al.(2025)Sun, Wu, Xu, Gao, Xu, Chen, Zhang, Ma, Zelek, and Li]{SAGonline}
Wentao Sun, Quanyun Wu, Hanqing Xu, Kyle Gao, Zhengsen Xu, Yiping Chen, Dedong Zhang, Lingfei Ma, John~S. Zelek, and Jonathan Li.
\newblock Sagonline: Segment any gaussians online.
\newblock \emph{arXiv preprint}, 2025.
\newblock arXiv:2508.08219.

\bibitem[Tancik et~al.(2020)Tancik, Srinivasan, Mildenhall, Fridovich-Keil, Raghavan, Singhal, Ramamoorthi, Barron, and Ng]{sam_pe}
Matthew Tancik, Pratul Srinivasan, Ben Mildenhall, Sara Fridovich-Keil, Nithin Raghavan, Utkarsh Singhal, Ravi Ramamoorthi, Jonathan Barron, and Ren Ng.
\newblock Fourier features let networks learn high frequency functions in low dimensional domains.
\newblock \emph{Advances in neural information processing systems}, 33:\penalty0 7537--7547, 2020.

\bibitem[Wang and Agapito(2024)]{wang2024spann3r}
Hengyi Wang and Lourdes Agapito.
\newblock 3d reconstruction with spatial memory.
\newblock \emph{arXiv preprint arXiv:2408.16061}, 2024.

\bibitem[Wang et~al.(2025{\natexlab{a}})Wang, Chen, Karaev, Vedaldi, Rupprecht, and Novotny]{wang2025vggt}
Jianyuan Wang, Minghao Chen, Nikita Karaev, Andrea Vedaldi, Christian Rupprecht, and David Novotny.
\newblock Vggt: Visual geometry grounded transformer.
\newblock In \emph{Proceedings of the IEEE/CVF Conference on Computer Vision and Pattern Recognition}, 2025{\natexlab{a}}.

\bibitem[Wang et~al.(2025{\natexlab{b}})Wang, Zhang, Holynski, Efros, and Kanazawa]{wang2025cut3r}
Qianqian Wang, Yifei Zhang, Aleksander Holynski, Alexei~A. Efros, and Angjoo Kanazawa.
\newblock Cut3r: Continuous updating transformer for 3d reconstruction.
\newblock In \emph{Proceedings of the IEEE/CVF Conference on Computer Vision and Pattern Recognition (CVPR)}, 2025{\natexlab{b}}.

\bibitem[Wang et~al.(2025{\natexlab{c}})Wang, Zhou, Zhu, Chang, Zhou, Li, Chen, Pang, Shen, and He]{wang2025pi3}
Yifan Wang, Jianjun Zhou, Haoyi Zhu, Wenzheng Chang, Yang Zhou, Zizun Li, Junyi Chen, Jiangmiao Pang, Chunhua Shen, and Tong He.
\newblock Scalable permutation-equivariant visual geometry learning.
\newblock \emph{arXiv preprint arXiv:2507.13347}, 2025{\natexlab{c}}.

\bibitem[Wang et~al.(2024)Wang, Leroy, Cabon, Chidlovskii, and Revaud]{wang2024dust3r}
Zian Wang, Vincent Leroy, Yohann Cabon, Boris Chidlovskii, and Jerome Revaud.
\newblock Dust3r: Geometric 3d vision made easy.
\newblock In \emph{Proceedings of the IEEE/CVF Conference on Computer Vision and Pattern Recognition (CVPR)}, pages 20697--20709, 2024.

\bibitem[Wu et~al.(2024)Wu, Jiang, Wang, Liu, Liu, Qiao, Ouyang, He, and Zhao]{wu2024ptv3}
Xiaoyang Wu, Li~Jiang, Peng-Shuai Wang, Zhijian Liu, Xihui Liu, Yu~Qiao, Wanli Ouyang, Tong He, and Hengshuang Zhao.
\newblock Point transformer v3: Simpler, faster, stronger.
\newblock In \emph{Proceedings of the IEEE/CVF Conference on Computer Vision and Pattern Recognition (CVPR)}, 2024.

\bibitem[Xu et~al.(2025)Xu, Yin, Qiu, Liu, Tong, and Han]{xu2025sampro3d}
Mutian Xu, Xingyilang Yin, Lingteng Qiu, Yang Liu, Xin Tong, and Xiaoguang Han.
\newblock Sampro3d: Locating sam prompts in 3d for zero-shot instance segmentation.
\newblock In \emph{International Conference on 3D Vision (3DV)}, 2025.

\bibitem[Yang et~al.(2025)Yang, Sax, Liang, Henaff, Tang, Cao, Chai, Meier, and Feiszli]{Yang_2025_Fast3R}
Jianing Yang, Alexander Sax, Kevin~J. Liang, Mikael Henaff, Hao Tang, Ang Cao, Joyce Chai, Franziska Meier, and Matt Feiszli.
\newblock Fast3r: Towards 3d reconstruction of 1000+ images in one forward pass.
\newblock In \emph{Proceedings of the IEEE/CVF Conference on Computer Vision and Pattern Recognition (CVPR)}, June 2025.

\bibitem[Yang et~al.(2023)Yang, Gao, Li, Gao, Wang, and Zheng]{yang2023track}
Jinyu Yang, Mingqi Gao, Zhe Li, Shang Gao, Fangjing Wang, and Feng Zheng.
\newblock Track anything: Segment anything meets videos.
\newblock \emph{arXiv preprint arXiv:2304.11968}, 2023.

\bibitem[Yeshwanth et~al.(2023)Yeshwanth, Liu, Nie{\ss}ner, and Dai]{yeshwanth2023scannetpp}
Chandan Yeshwanth, Yueh-Cheng Liu, Matthias Nie{\ss}ner, and Angela Dai.
\newblock Scannet++: A high-fidelity dataset of 3d indoor scenes.
\newblock \emph{arXiv preprint arXiv:2308.11417}, 2023.
\newblock URL \url{https://arxiv.org/abs/2308.11417}.

\bibitem[Ying et~al.(2024)Ying, Yin, Zhang, Wang, Yu, Huang, and Fang]{OmniSeg3D}
Haiyang Ying, Yixuan Yin, Jinzhi Zhang, Fan Wang, Tao Yu, Ruqi Huang, and Lu~Fang.
\newblock Omniseg3d: Omniversal 3d segmentation via hierarchical contrastive learning.
\newblock In \emph{Proceedings of the IEEE/CVF Conference on Computer Vision and Pattern Recognition (CVPR)}, 2024.

\bibitem[Zhang et~al.(2025)Zhang, Wang, Xu, Xue, Rupprecht, Zhou, Shen, and Wetzstein]{Zhang_2025_FLARE}
Shangzhan Zhang, Jianyuan Wang, Yinghao Xu, Nan Xue, Christian Rupprecht, Xiaowei Zhou, Yujun Shen, and Gordon Wetzstein.
\newblock Flare: Feed-forward geometry, appearance and camera estimation from uncalibrated sparse views.
\newblock In \emph{Proceedings of the IEEE/CVF Conference on Computer Vision and Pattern Recognition (CVPR)}, June 2025.

\end{thebibliography}

\end{document}